\pdfoutput=1
\documentclass{bmvc2k}

\title{Efficient Feature Extraction for High-resolution Video Frame Interpolation} 

\addauthor{Moritz Nottebaum}{moritz.nottebaum@stud.tu-darmstadt.de}{1}
\addauthor{Stefan Roth}{stefan.roth@visinf.tu-darmstadt.de}{1,2}
\addauthor{Simone Schaub-Meyer}{simone.schaub@visinf.tu-darmstadt.de}{1,2}

\addinstitution{
Department of Computer Science\\
TU Darmstadt
}
\addinstitution{
hessian.AI
}

\runninghead{Nottebaum \etal{}}{Feature Extraction for Video Frame Interpolation}

\def\eg{\emph{e.g}\bmvaOneDot}

\def\etal{\emph{et al}\bmvaOneDot}

\usepackage[utf8]{inputenc} 
\usepackage[T1]{fontenc}    
\usepackage{hyperref}       
\usepackage{url}            
\usepackage{booktabs}       
\usepackage{amsfonts}       
\usepackage{nicefrac}       
\usepackage{microtype}      
\usepackage{xcolor}         
\usepackage{multibib}
\newcites{latex}{References}
\usepackage[page]{appendix}

\usepackage{graphicx}
\usepackage{amsmath}
\usepackage{bbding}
\usepackage{tabularx}
\usepackage{booktabs}
\usepackage{multirow}
\usepackage{newtxtt}
\usepackage{etoolbox,siunitx}
\robustify\bfseries
\usepackage{wrapfig}

\usepackage[capitalize]{cleveref}
\crefname{section}{Sec.}{Section}
\crefname{table}{Tab.}{Tabs.}

\usepackage{pifont} 
\newcommand{\cmark}{\ding{51}}%
\newcommand{\xmark}{\ding{55}}%

\usepackage{xspace}
\newcommand{\ie}{\emph{i.\thinspace{}e.}\@\xspace}

\newcommand{\wrt}{\emph{w.\thinspace{}r.\thinspace{}t.}\@\xspace}

\newcommand\blfootnote[1]{%
  \begingroup
  \renewcommand\thefootnote{}\footnote{#1}%
  \addtocounter{footnote}{-1}%
  \endgroup
}

\newcommand{\fLDR}{fLDR}
\newcommand{\papername}{\fLDR-Net}
\newcommand{\fzo}{\ensuremath{F_{0 \rightarrow 1}}}
\newcommand{\foz}{\ensuremath{F_{1 \rightarrow 0}}}
\newcommand{\mftz}{F_{t \rightarrow 0}}
\newcommand{\ftz}{$\mftz$}
\newcommand{\mfto}{F_{t \rightarrow 1}}
\newcommand{\fto}{$\mfto$}
\newcommand{\mfzt}{F_{0 \rightarrow t}}
\newcommand{\mfot}{F_{1 \rightarrow t}}
\newcommand{\mIzt}{I_{0 \rightarrow t}}
\newcommand{\mIot}{I_{1 \rightarrow t}}
\newcommand{\mItz}{I_{t \leftarrow 0}}
\newcommand{\mIto}{I_{t \leftarrow 1}}
\newcommand{\softsplat}[2]{\overrightarrow{\sigma}(#1, \, #2)}

\begin{document}

\maketitle

\begin{abstract}
Most deep learning methods for video frame interpolation consist of three main components: feature extraction, motion estimation, and image synthesis. 
Existing approaches are mainly distinguishable in terms of how these modules are designed. 
However, when interpolating high-resolution images, \eg~at 4K, the design choices for achieving high accuracy within reasonable memory requirements are limited.
The feature extraction layers help to compress the input and extract relevant information for the latter stages, such as motion estimation. However, these layers are often costly in parameters, computation time, and memory.
We show how ideas from dimensionality reduction combined with a lightweight optimization can be used to compress the input representation while keeping the extracted information suitable for frame interpolation. 
Further, we require neither a pretrained flow network 
nor a synthesis network, additionally reducing the number of trainable parameters and required memory.
When evaluating on three 4K benchmarks, we achieve state-of-the-art image quality among the methods without pretrained flow while having the lowest network complexity and memory requirements overall.
\blfootnote{Code and additional resources (test set and supplemental material) are available at \url{ https://github.com/visinf/fldr-vfi}}
\end{abstract}

\section{Introduction}
\label{sec:intro}
Video frame interpolation (VFI) is one of the classic problems in video processing. Generating intermediate frames in a video sequence can be used for various applications,  
\eg, video compression \citep{Djelouah:2019:NIF, Wu:2018:VCI}, video editing \citep{Meyer:2018:DVC},  animation \citep{Briedis:2021:NFI, Li:2021:DAV}, and event cameras \citep{Tulyakov:2021:TLE, Tulyakov:2022:TLE}.

Deep learning-based approaches for VFI~\citep{Liu:2017:VFS, Sim:2021:XVF,Jiang:2018:SSH, Park:2021:ABM} commonly consist of three modules: feature extraction, motion estimation, and image synthesis, which can be realized and combined in various forms. The higher the input image resolution is, however, the more important an efficient feature representation becomes to allow the subsequent modules to handle the large input space within reasonable memory constraints. Convolutional neural networks (CNNs) can reduce the spatial dimensions with an encoder network. This still requires initially applying the convolutional filters at the full image resolution, however.
We suggest instead to first compress the images with a block-wise transformation inspired by linear dimensionality reduction methods such as principal component analysis (PCA) \citep{Pearson:1901:PCA}. PCA is known to efficiently represent image data, but is not directly suitable as a representation for neural networks as it projects each input image to a different low-dimensional space. 
We propose a method that finetunes an initial block-based PCA basis end-to-end for video frame interpolation. This has two advantages: First, it results in one general projection space for all images and resolutions. Second, it allows us to optimize the representation for the targeted task of video frame interpolation.

Our main contribution is a deep learning architecture for video frame interpolation that first applies linear dimensionality reduction to efficiently compress and represent images. Our approach is carefully designed to learn from the projected data as well as avoids a memory and computation intensive synthesis network in favor of a learned weighting of the (warped) input images. As a result, the overall complexity of the network architecture can be significantly reduced, which makes it particularly suitable for high-resolution videos and limited hardware resources. To evaluate the generalization of our approach to various video scenes, we introduce a novel 4K test set, with an order of magnitude more video scenes than existing ones. We achieve highly competitive results and even outperform existing methods for larger motions across various benchmarks with only a fraction of the network complexity. 

\section{Related Work}
\label{sec:related_work}
\paragraph{Feature extraction.}
Most deep learning-based VFI methods encode the input with 
CNNs due to their general suitability for feature representation~\citep{Simonyan:2015:VDC, Krizhevsky:2012:INC}. However, for compressing images and videos, algorithms used in standard codecs~\citep{Wallace:1991:JPE, Taubman:2002:JPE, Wiegand:2003:OHV} still rely on mathematical transformations like the discrete cosine transform (DCT). 
The advantage of well defined mathematical transformations like DCT, PCA, and DFT (discrete Fourier transformation) is that they are comparably cheap to compute while also entailing a good compression ratio. Some exemplary work showed their use in neural networks in different scenarios, \eg, DFT~\citep{Riad:2022:LSC,Meyer:2018:PNV} and PCA~\citep{Benkaddour:2017:FEC}.
The advantage of CNNs is their task-specific adaptability, while PCA dimensionality reduction only adapts to the input data, not the task. Here, we show how we can still use the benefits of dimensionality reduction in a neural network architecture.

\paragraph{Video frame interpolation.}
Generating intermediate frames between two frames requires some understanding of the temporal relationship between the input frames. We can generally categorize VFI methods in terms of whether they estimate the motion explicitly, \eg optical flow, or implicitly.
VFI methods representing the motion implicitly either directly estimate the intermediate frame~\citep{Long:2016:LIM}, use phase-based~\citep{Meyer:2015:PBF, Meyer:2018:PNV} or a kernel-based 
\citep{Niklaus:2017:VFI, Niklaus:2017:VFIAS, Cheng:2020:VFI} representation, or apply PixelShuffle~\citep{Choi:2020:CAA}. However, most of them are limited to generating a single intermediate frame. Kernel-based methods are further limited, due to the kernel size, in the amount of pixel displacement they can handle. This makes them unsuitable for high-resolution imagery.

Traditionally, explicit methods for VFI combine optical flow estimation \citep{Baker:2011:DBE,Sun:2010:SOF} and correspondence-based image warping \citep{Baker:2011:DBE}. As a result, they heavily depend on the quality of the optical flow. Thus, deep learning-based methods using flow estimation often not only estimate the bi-directional flow between the input images but further refine the intermediate flow vectors~\citep{Jiang:2018:SSH, Sim:2021:XVF} or the output with a synthesis network~\citep{Niklaus:2018:CAS} to address inaccuracies.
Two recent advances have enabled the current state-of-the-art in frame interpolation: On the one hand, pretrained flow estimation modules~\citep{Sun:2018:PWC, Teed:2020:RAF, Xu:2022:GMF} became more powerful and can be used as a starting point; on the other hand, using forward warping instead of backward warping allows to directly compute the intermediate flow vectors without approximating them~\citep{Niklaus:2020:SSV,Hu:2022:MMS}. However, their performance heavily depends on the pretrained flow network used as well as on the image resolution of the input~\citep{Hu:2022:MMS}. M2M-PWC~\citep{Hu:2022:MMS}, for example, requires bilinear downscaling of 4K images for optimal performance. Furthermore, such models have many trainable parameters and require a lot of memory.

Applying VFI techniques successfully and efficiently to high-resolution data remains a
challenging task. 
Most successful high-resolution models have in common that they do not compute the flow via an expensive cost volume \citep[\eg,][]{Huang:2020:RIF} and use a multi-scale approach, incorporating some kind of weight sharing between the different scales of the input images \citep[\eg,][]{Sim:2021:XVF, Park:2021:ABM,Reda:2022:FIL}. 
We take these ideas a step further and combine the shared multi-resolution approach with a preceding image compression based on linear dimensionality reduction.

\section{Dimensionality Reduction for Feature Extraction}
\label{sec:PCA}
As our approach to obtaining a compact image representation is inspired by principal component analysis (PCA) \cite{Pearson:1901:PCA}, let us first recap its basics. PCA describes the process of computing principal components from the data. 
The principal components represent an orthonormal basis of the zero-centered data. PCA can be used to perform a change of basis of the data, where the first basis vector maximizes the variance of the projected data, the second accounts for the largest remaining variability, and so on.
Because of this characteristic, the PCA basis is commonly used for dimensionality reduction by projecting each data point onto only the first few principal components. As a result, the lower-dimensional, projected data preserves as much of the data's variation as possible with a linear transformation.

Instead of extracting a compact feature representation with a neural network directly from the high-resolution input image, we here argue that dimensionality reduction can be used to first compress the image data. This reduces the memory footprint of the input data significantly without resorting to downscaling. There are two main advantages of combining explicit dimensionality reduction with neural networks: \emph{(i)} Using a compressed representation of the input reduces the memory requirements of the subsequent neural network modules. \emph{(ii)} Due to the compressed representation also the model capacity of the subsequent neural network modules can be reduced, decreasing the number of trainable parameters.

\paragraph{Block-based PCA.}
Similar to \citep{Taur:1996:MIC}, we use block-based PCA to compress a single image. We split an image $I \in \mathbb{R}^{H\times W\times 3}$ into $d\times d$ blocks, where each block, vectorized, represents a data point $x_i \in \mathbb{R}^{d^2}$, which are concatenated into a matrix as $X=[x_1, \dots, x_N] \in \mathbb{R}^{d^2\times N}$. The number of blocks $N$ depends on the image resolution, \ie, $N=3HW/d^2$. After subtracting the mean $\Bar{x}=\nicefrac{1}{N} \sum_i x_i$ from $X$, resulting in $\Bar{X}$, we compute the covariance matrix $C \in \mathbb{R}^{d^2\times d^2}$ of $\Bar{X}$ as $C=\nicefrac{1}{N-1}\cdot\Bar{X}\Bar{X}^T$. We perform an eigendecomposition of $C$ and take the first $k$ eigenvectors $u_i$ as our projection axes, \ie, $U_k=[u_1, \dots , u_k] \in \mathbb{R}^{d^2\times k}$. We compute the projected data as $\tilde{X}=U_k^T\Bar{X}, \, \tilde{X} \in \mathbb{R}^{k\times N}$. Reshaping $\tilde{X}$ back yields the compressed image representation $\tilde{I} \in \mathbb{R}^{H/d\times W/d \times 3k}$ with a compression ratio of $r=k/d^2$ \wrt the original image $I$. 

Simply applying block-based PCA to each image is computationally not ideal as the principal components have to be recomputed every time. Further, the projected data is not directly suitable as input to a neural network. First, the data range can be very large and second, representing each image in a different projected space makes it difficult for the network to learn from it. We address these challenges explicitly in our framework design. 

\section{\papername{} for Video Frame Interpolation}
\label{sec:method}
Given two images, $I_0$ and $I_1$, the goal of video frame interpolation is to generate several intermediate images $I_t$ for $t \in (0,1)$. 
The focus of our method is to propose a lightweight framework in terms of memory and the number of trainable parameters that is especially suitable for high-resolution images, in our case 4K, and limited hardware resources.
We achieve this by compressing the input images using ideas from dimensionality reduction and projecting the data into a low-dimensional space. However, it is not trivial how to use this different representation efficiently in a neural network.
Following the recap of the basics of principal component analysis PCA in~\cref{sec:PCA}, we now explain how to use a block-based linear representation successfully in a neural network, as well as our overall framework design to further reduce the computational overhead.

\begin{figure}[t]
     \centering
     \includegraphics[width=0.9\textwidth]{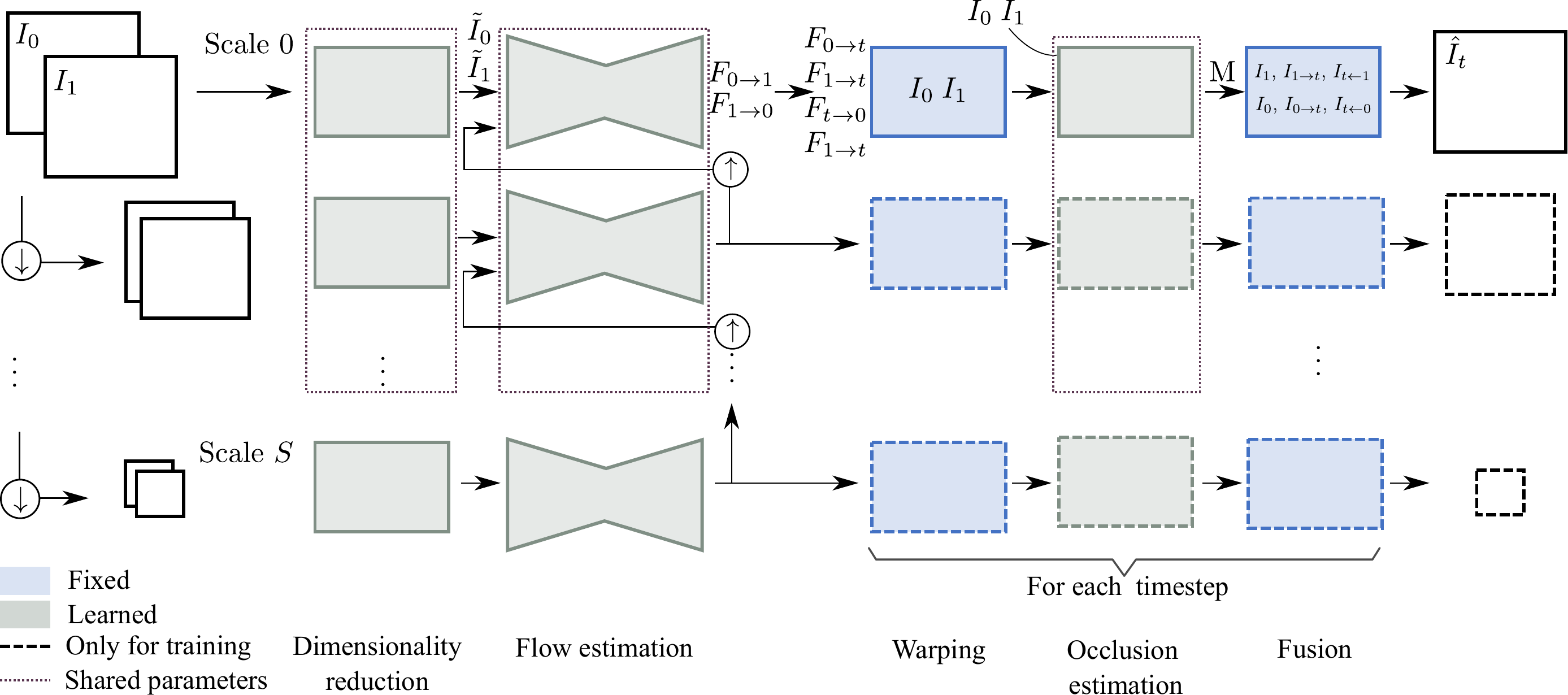}
    \vspace{-.5em}
     \caption{\textbf{\papername{} architecture.} Given two images, $I_0$ and $I_1$, we generate the intermediate image $\hat{I}_t$ at time $t$ by applying our finetuned dimensionality reduction followed by flow estimation, image warping (forward and backward), and estimation of the weighting map $M$ for fusing the warped images, giving the final result. Our method is very lightweight in terms of trainable parameters as the parameters are shared across scales; warping and fusion are fixed, non-learnable computations. The synthesis of $\hat{I}_t^s$ for $s>0$ is only used during training.
     }
\label{fig:overview}
\end{figure}

\paragraph{Overview.}
\label{sec:architecture}
\cref{fig:overview} shows our proposed framework. It consists of three trainable modules, namely the finetuned linear dimensionality reduction (\fLDR{}), flow estimation, and occlusion estimation. The image synthesis part at each time step $t$ is very lightweight in terms of trainable network modules and memory. We do not perform any learned refinement of the estimated flow vectors or image synthesis besides the occlusion estimation. 
In the following, we describe each part in detail. To simplify the notation, we focus on scale level $s=0$. If not mentioned otherwise, the other scales are carried out in the same way except that the input image is first bilinearly downscaled by a factor of $2$ from the previous level.

\paragraph{Finetuned linear dimensionality reduction (fLDR).}
We apply dimensionality reduction of the input images inspired by block-based PCA as described in \cref{sec:PCA}. However, we cannot apply this to each image $I$ separately as this would lead to different, data-dependent low-dimensional projection spaces, unfavorable for neural networks. Additionally, the projection spaces should be consistent among the scales, in order to enable parameter sharing. This requires a single set of projection vectors suitable for all images at all scales. We achieve this by using the $k$ first block-based PCA projection vectors $U_k$ of one random sample image as initialization and treating the eigenvectors and the mean vector of the
computed initial PCA decomposition as trainable parameters, optimized end-to-end for the task of frame interpolation. 
We call this procedure finetuned block-based linear dimensionality reduction, short \fLDR{}, yielding the compressed image $\tilde{I}$. 
Before further processing $\tilde{I}$, we divide each projected data point by the mean of its absolute values along the principal direction dimensions and normalize the whole matrix to values in the range of $[-1,1]$ suitable for neural networks.

\paragraph{Flow estimation.}
Existing flow estimation modules often estimate flow only at a downsampled input resolution, \eg $\nicefrac{1}{4}$ (\citep{Sun:2018:PWC, Sim:2021:XVF}) or $\nicefrac{1}{8}$ (\citep{Teed:2020:RAF}), for computational reasons. Because the input to the flow estimation is in our case already a compressed image representation obtained from the \fLDR{} module, we estimate the flow at the same resolution as the compressed input representation, \ie, at $\nicefrac{1}{d}$ resolution. We use the high-level idea of XVFI~\citep{Sim:2021:XVF} of resharing the parameters across scales
for efficient multi-scale optical flow estimation without a pretrained flow network. 
We also estimate the flow at different scales $s=0,\ldots, S$ 
with shared parameters, except for the lowest level $s=S$. 
This allows to increase the number of scales at test time, when the input image has a significantly larger resolution than at training time (\eg, 4K \emph{vs.}\ training patch size). 
Contrary to XVFI~\citep{Sim:2021:XVF}, we can further significantly simplify the flow estimation module and only need one encoder-decoder network instead
of two to predict the intermediate flow fields. The reasons are our compressed image representation as input as well as changes in the pipeline, mainly using forward warping instead of backward warping.

As illustrated in \cref{fig:overview},  we predict \fzo{} and \foz{} at each scale . The flow is estimated from the compressed image representation $\tilde{I}$, which first gets passed through two convolutional layers. On all levels except $s=S$, the feature representation is concatenated with the bilinearly upscaled flow estimation from the previous level. Architectural network details are provided in the supplemental material.

To compute the intermediate image $I_t$, we need the flow from and to the intermediate time step $t$. To compute the forward flow, we can simply scale it accordingly:
\begin{align}
    \mfzt = t \fzo \text{\quad and \quad}
    \mfot = (1-t) \foz \; .
\end{align}

Computing the backward flow, \ftz\ and \fto, is only possible in an approximated manner~\citep{Jiang:2018:SSH}. XVFI~\citep{Sim:2021:XVF} uses a second neural network for this, while we opt for a direct, non-trainable approximation. In order to obtain \ftz, we can take the flow \foz, adjusting the magnitude by multiplying it with $t$ and warping it to the position from which it points at time $t$. We do the latter by backward warping $t\foz$ with the flow \fzo, scaled with $1-t$, approximating the starting point of the flow vector at time $t$. $\mfto{}$ is approximated analogously as
\begin{align}
\label{eqn:ftz}
\mftz{} \approx \overleftarrow{\omega} ( t\foz ,\, (1-t)  \fzo ) \text{\quad and \quad}
\mfto{} \approx \overleftarrow{\omega} ((1-t)\fzo ,\, t  \foz) \; ,
\end{align}
where $\overleftarrow{\omega}(\cdot, \cdot)$ describes the backward warping function.

\paragraph{Image warping.}
Before we can use the computed flow vectors for warping, we upscale them bilinearly to the non-compressed image resolution at each scale.
We can then warp the input images, $I_0$ and $I_1$, to the intermediate time step $t$. We do this by forward warping them with softmax splatting~\cite{Niklaus:2020:SSV}, noted as $\overrightarrow{\sigma}$, with occlusion estimation~\citep{Baker:2011:DBE} as importance metric:
\begin{align}
\mIzt = \softsplat{I_0}{\mfzt} \text{\quad and \quad}
\mIot = \softsplat{I_1}{\mfot} \; .
\end{align}
The advantage of forward splatting is that we can directly use the intermediate flow vectors $\mfzt{}$ and $\mfot$. However, forward splatting can lead to small holes due to occlusions and divergent flow vectors. Backward warping, on the other hand, would give a dense result, but requires an approximation of the backward flow. Video frame interpolation methods that use backward warping often use an additional network to refine $\mftz{}$ and $\mfto{}$, \eg, XVFI~\citep{Sim:2021:XVF} for better accuracy. In our case, we only use the backward images as additional input to compute the occlusion mask. Therefore, the directly computed flow vectors from \cref{eqn:ftz} are enough and no further learned refinement is needed. The backward warping gives us two additional images at time instant $t$:
\begin{align}
\mItz = \overleftarrow{\omega}(I_0, \, \mftz) \text{\quad and \quad}
\mIto = \overleftarrow{\omega}(I_1, \, \mfto) \; .
\end{align}

\paragraph{Occlusion estimation.}
In order to accurately merge the different warped images at time step $t$, we propose a simple occlusion estimation network to estimate a weighting map $M \in \mathbb{R}^{H\times W \times c}$, where $c$ corresponds to the number of images to merge. The input to the occlusion estimation are the warped images as well as the input images.  In the last layer of the estimation network, we apply a softmax along the last dimension, yielding a probability distribution for every output pixel. We include temperature scaling~\citep{Guo:2017:CMN} of the softmax function to better calibrate the distribution.
The details of the temperature scaling and the network architecture can be found in the supplemental material.

\paragraph{Image synthesis.}
We omit a dedicated image synthesis network to predict $\hat{I}_t$. Instead, we directly fuse the warped images by weighting them with the predicted occlusion map $M$ (by pixelwise multiplication $\odot$) as well as the temporal distance from the input images:

\begin{align}
\hat{I}_{t} = \frac{\sum_{i\in{0,1}} \left[(i \cdot t) + (1-i)(1-t)\right] \cdot \left( M_{t \leftarrow i} \odot  I_{t \leftarrow i} + M_{i \rightarrow t} \odot I_{i \rightarrow t} + M_{i} \odot I_i\right)}
{(1-t) \cdot \left(M_{t \leftarrow 0} + M_{0 \rightarrow t} + M_{0}\right) + t \cdot \left(M_{t \leftarrow 1} + M_{1 \rightarrow t} +  M_{1}\right)} \; .
\label{eq:image_synthesis_new}
\end{align}

\paragraph{Loss functions.}
We train our model with a similar loss function as in XVFI~\citep{Sim:2021:XVF}, consisting of an image reconstruction loss, $\mathcal{L}_{recon}$, at all levels and an edge-aware smoothness loss, $\mathcal{L}_{smooth}$, 
on the estimated flow vectors \fzo\ and \foz\ at the finest scale $s=0$. We add an additional term to supervise the flow estimation at the highest resolution by computing the warping error, $\mathcal{L}_{warp}$, yielding the total loss
\begin{align}
\label{eqn:total_loss}
    \mathcal{L}_{total} &= \mathcal{L}_{recon} + \lambda_{smooth} \cdot \mathcal{L}_{smooth} + \lambda_{warp} \cdot \mathcal{L}_{warp} \; ,
\end{align}
where $\lambda_{(\cdot)}$ define the weighting factors.
Details are given in the supplemental.

\section{Experiments}
\label{sec:experiments}
\subsection{Setup}
\paragraph{Training.}
We train our method on X-Train~\citep{Sim:2021:XVF}, mainly following the training procedure of XVFI~\citep{Sim:2021:XVF}. 
X-Train consists of 4,408 clips of
$768\times 768$ images cropped from the original 4K frames. Each clip consists of 65 consecutive frames. Following~\citep{Sim:2021:XVF}, we also select random triplets from the sequences, at most 32 frames apart, and randomly select training patches of $512\times 512$. We train on one Nvidia 3080Ti (12GB) GPU with a batch size of 8 for 200 epochs. We use the Adam~\citep{Kingma:2015:AMS} optimizer and an initial learning rate of $10^{-4}$, which is decreased by 0.25 after each 50 epochs, starting after 70 epochs. The optimization of the temperature parameter $T$ \citep{Guo:2017:CMN} is done with a learning rate of $10^{-3}$. 
Because the projection vectors are very sensitive to changes, we use a learning rate of $10^{-5}$ for the linear dimensionality reduction layer. We use $S=3$ for training. For evaluation, we take the best checkpoint based on the validation set. The full list of hyperparameters is given in the supplemental.

\paragraph{Testing.}
We use $S=5$ during testing, \ie 2 scale levels more than during training to handle the high-resolution images. 
For evaluation, we focus on datasets consisting of 4K video frames. The two commonly used 4K datasets in frame interpolation are the videos from Xiph\footnote{from \url{xiph.org}} as described in \citep{Niklaus:2020:SSV}, as well as X-Test from~\citep{Sim:2021:XVF}. However, both of them have limited variability. X-Test only has 15 image pairs, each with 7 intermediate frames. The selected clips have a lot of camera motion leading to large, dominating flow magnitudes. Xiph, on the other hand, has only 8 videos, with much less camera motion and mainly object motion, resulting in most of the flow vectors having a small magnitude. This is summarized in~\cref{table:dataset}.  To have more variability, we selected additional evaluation scenes from the Inter4K~\citep{Stergiou:2021:APE} test set. The original test set consists of 100 videos with up to 60 fps and 300 frames per video. We only take the videos with 60 fps and split them into their different scenes by thresholding the pixel difference between consecutive frames. We create two different versions to evaluate different degrees of motion. For Inter4K-S, we take the first 9 frames of each scene, and for Inter4K-L the first 17, respectively. 
Both test sets analyze 8$\times$ temporal interpolation with 7 intermediate frames, meaning for 
 Inter4K-L every second intermediate frame is skipped to simulate larger motion.
\setlength{\tabcolsep}{2pt}
\begin{table}[t]
\footnotesize
\begin{center}
\begin{tabularx}{\linewidth}{@{}l*7{>{\centering\arraybackslash}X}@{}}
\toprule
 & \multirow{3}{*}{\shortstack{Interpolation\\ factor}} & \multirow{3}{*}{\shortstack{\# clips\\ \;}} & \multirow{3}{*}{\shortstack{\# eval.\\ frames $/$ clip}} & \multirow{3}{*}{\shortstack{\# total eval. \\frames}} & \multicolumn{3}{c}{PWC-Net~\citep{Sun:2018:PWC}} \\
\cmidrule{6-8}
Dataset &&&&& $25_\mathit{th}$ & $50_\mathit{th}$ & $75_\mathit{th}$ \\
\midrule
Xiph-4K & 2$\times$ & 8 & 49 & 392  & 9.1 & 14.1 & 25.0  \\ 
X-Test~\citep{Sim:2021:XVF} &  8$\times$ & 15 & 7 & 105   & 23.9 & 81.9 &  138.5 \\
Inter4K-S & 8$\times$ & 144 & 7 & 1008 & 7.6 & 30.9 & 102.0 \\
Inter4K-L & 8$\times$ & 133 & 7 & 931  & 13.4 & 56.7 & 162.5 \\
\bottomrule
\end{tabularx}
\end{center}
\caption{\textbf{Overview of used datasets.} Comparison of our newly proposed test sets based on Inter4K~\citep{Stergiou:2021:APE} to existing datasets. Inter4K-S and Inter4K-L cover more scenes and Inter4K-L has statistically larger motion. The percentiles of the flow magnitudes (last 3 columns) are computed from the estimated flow~\citep{Sun:2018:PWC} between the input frames. }
\label{table:dataset}
\vspace{-.5em}
\end{table}

\setlength{\tabcolsep}{4pt}
\begin{table}[t]
\footnotesize
\begin{center}
\begin{tabularx}{\linewidth}{@{}l*9{>{\centering\arraybackslash}X}@{}}
\toprule
& Pretrained flow & \# Param. (Mill.) & Memory (for 4K) & Training dataset & Xiph-4K
& X-Test 
& Inter4K-S & Inter4K-L & Inference (in s/f)\\
\midrule
SoftSplat~\cite{Niklaus:2020:SSV} & \cmark    & 7.7  & --  & Vimeo90K   &   34.2$^\ast$\citep{niklaus2022splatting}   &  25.48$^\ast$\citep{Hu:2022:MMS} & OOM & OOM & --\\
M2M-PWC~\cite{Hu:2022:MMS} & \cmark  &  7.6  &  10 GB    &  Vimeo90K  &   \textbf{34.88}   &   \textbf{30.81}   & \underline{29.22}   & \underline{24.87} & \textbf{0.21}\\
M2M-DIS~\cite{Hu:2022:MMS} & \cmark  & --  &  --   &  Vimeo90K  &       --  &  30.18$^\ast$\citep{Hu:2022:MMS}& --  &  -- & --\\
\midrule
ABME~\cite{Park:2021:ABM} &    \xmark   & 18.1$^\ast$\citep{Park:2021:ABM} & -- &  Vimeo90K &     OOM  &  30.16$^\ast$\citep{Park:2021:ABM}  &  OOM  &  OOM & --\\
RIFE$_m$~\cite{Huang:2020:RIF}  &  \xmark   & 9.8   &  \underline{6.8} GB    &   Vimeo90K     &  \underline{34.80}   &  26.80  & 28.37   &  24.40 & \underline{0.40}\\
RIFE$_{m}^\diamond$~\cite{Huang:2020:RIF} &  \xmark   & 9.8   &   \underline{6.8} GB     &  X-Train   &  34.00  & 28.06  & 28.36  &  24.47 & \underline{0.40}\\
XVFI$^\diamond$ ~\cite{Sim:2021:XVF} &  \xmark       &  \underline{5.5} &  >12 GB   &  X-Train &    34.04  &  30.34 & 28.82  & 24.62 & --\\
Ours & \xmark   &  \textbf{0.9}  &    \textbf{4.6 }GB   &   X-Train     &  34.16  &  \underline{30.45}   & \textbf{29.29 }  &  \textbf{25.16} & 0.51\\
\bottomrule
\end{tabularx}
\end{center}
    \vspace{-.5em}
\caption{\textbf{Quantitative results} on 4K images in terms of PSNR. Numbers with $\ast$ are taken from the given references; $\diamond$ denotes retraining. The methods have been trained on either Vimeo90K~\citep{Xue:2019:VET} or X-Train~\citep{Sim:2021:XVF}. The inference time has been measured on a Nvidia 3080Ti (12GB) GPU. The best and 2nd best values are in \textbf{bold} and \underline{underlined}, respectively. An evaluation with SSIM~\citep{Wang:2004:IQA} and LPIPS~\citep{Zhang:2018:UED} is given in the supplemental.}
\label{table:results_4K}
\vspace{-1em}
\end{table}

\subsection{Comparison with the state of the art}
We compare to SoftSplat~\citep{Niklaus:2020:SSV} and M2M~\citep{Hu:2022:MMS}, which both rely on pretrained flow and forward splatting, as well as to ABME~\citep{Park:2021:ABM}, RIFE~\citep{Huang:2020:RIF}, and XVFI~\citep{Sim:2021:XVF}, which do not use a pretrained flow network.
All methods support interpolation at arbitrary time steps. 

Whenever possible, we used the original code and checkpoints from the authors. 
The performance of M2M~\citep{Hu:2022:MMS} heavily depends on the 
input image resolution to the flow network, \eg, M2M-PWC~\cite{Hu:2022:MMS} performs best for 4K input when images are first downscaled by a factor of 8 or 16, while M2M-DIS is less sensitive, and performs best at the original input. For M2M-DIS, we report the numbers from the original paper~\citep{Hu:2022:MMS} as the code is not online. For M2M-PWC, we test different scale factors for each dataset, reporting the results for the best option. For RIFE~\citep{Huang:2020:RIF}, we use the RIFE$_{m}$ model, which supports arbitrary interpolation. We also retrain it on X-Train with an increased patch size of $512 \times 512$, as we observed this to slightly improve the results.
We evaluate the models on a Nvidia 3080 Ti (12GB) GPU. With the available memory, we were not able the evaluate SoftSplat~\citep{Niklaus:2020:SSV} and ABME~\citep{Park:2021:ABM} on 4K images and hence report numbers from the referenced papers instead. 

\cref{table:results_4K} gives a quantitative comparison. We achieve state-of-the-art accuracy for most of the datasets among the methods without pretrained flow. A powerful pretrained flow network can occasionally help, but the results are heavily dependent on the type of flow network and amount of motion. Our approach yields consistently high PSNR values, especially for the datasets with larger motion, despite having by far the fewest trainable parameters as well as the lowest memory requirements. In comparison to XVFI~\citep{Sim:2021:XVF}, we can see how our conceptual contributions lead to a significant reduction of memory and parameters, yet an increase in PSNR of up to $0.54$dB. 
Without specifically optimizing for inference time, we can achieve competitive runtimes comparable to most of the other methods (M2M-PWC downscales the input image first by a factor of $\nicefrac{1}{16}$, leading to faster inference) without the need for large, high-end GPUs.
Qualitative visual results are shown in \cref{fig:result_visual} and in the supplemental video. 

\begin{figure}[t]
\begin{minipage}[t][0.42\textwidth][b]{0.48\textwidth}
     \centering
     \includegraphics[width=0.9\linewidth]{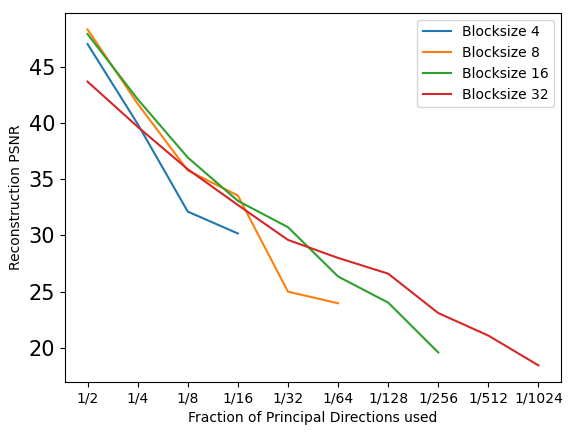}
     \caption{Reconstruction acc. for different block sizes $d$ \wrt the compression ratio $\nicefrac{1}{r}$.
     }    
 \label{fig:reconstruction_error}
\end{minipage}%
\hfill
\begin{minipage}[t][0.42\textwidth][b]{0.45\textwidth}
    \centering
    \subfigure[Original PCA]{
    \begin{minipage}[t]{0.9\linewidth}
        \centering
        \includegraphics[width=0.48\textwidth]{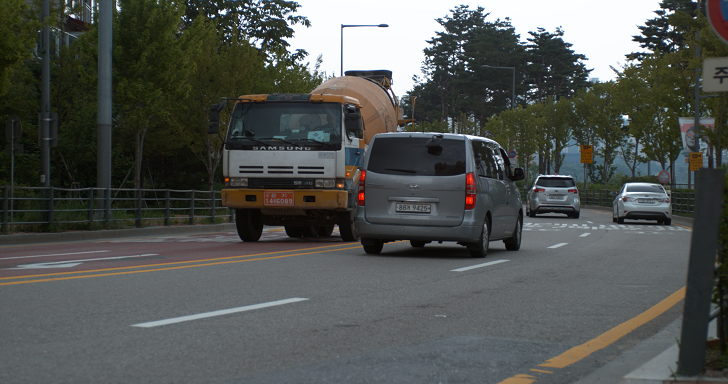}
        \includegraphics[width=0.48\textwidth]{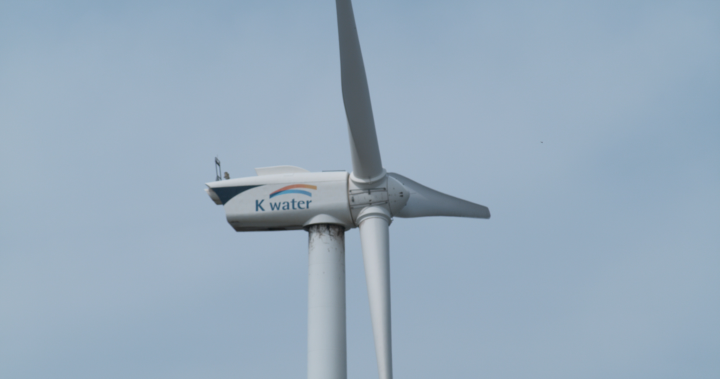}
    \end{minipage}
    \label{fig:originalPCA}
    }
    \subfigure[With finetuning]{
    \begin{minipage}[t]{0.9\linewidth}
        \centering
        \includegraphics[width=0.48\textwidth]{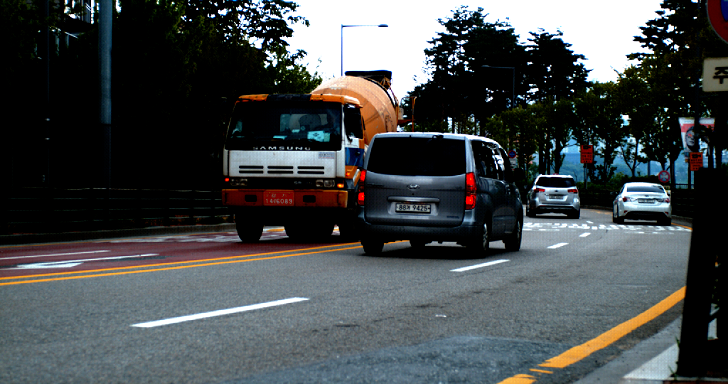}
        \includegraphics[width=0.48\textwidth]{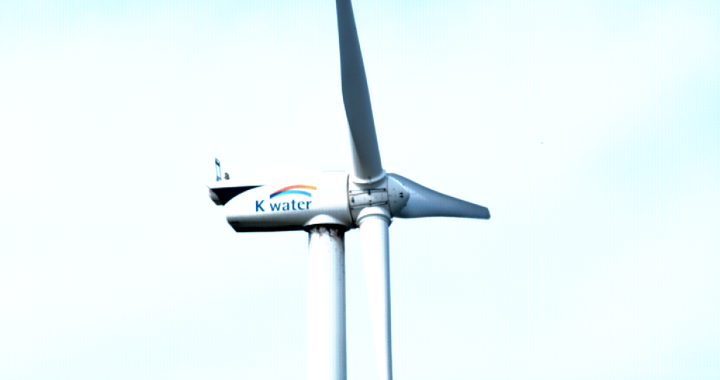}
        \label{fig:finetuned}
    \end{minipage}
    }
    \caption{Image reconstruction with a $\nicefrac{1}{4}$ of the projection directions, with and without finetuneing them.
    }
    \label{fig:comparison_features}
\end{minipage}
\vspace{-1em}
\end{figure}
\setlength{\tabcolsep}{4pt}

\setlength{\tabcolsep}{4pt}
\begin{table}[t]
\footnotesize
\begin{center}
\begin{tabularx}{\linewidth}{@{}l*6{>{\centering\arraybackslash}X}@{}}
\toprule
& fDLR & Flow estim. & Warping & Occl. estim. & Fusion & Full model\\
\midrule
Memory (GB) & 3.5 & 2.7  & 4.6  & 4.0  & 4.6 & 4.6 \\
\# Parameters & 1k & 733k & 1  & 194k & 0 & 0.9M\\
Inference time (ms) & 30  &  22  &  410  &  44  & 3.1 & 510 \\
\bottomrule
\end{tabularx}
\end{center}
    \vspace{-0.5em}
\caption{\textbf{Analysis of the computational resources.} Breakdown of the individual modules of our framework for performing inference on a 4K frame.}
\label{table:componenents}
\end{table}

\begin{table}[t]
\footnotesize
\begin{center}
\begin{tabularx}{\linewidth}{@{}l*6{>{\centering\arraybackslash}X}@{}}
\toprule
& \# Param. (in Mill.) & Memory (for 4K) & Xiph-4K & X-Test & Inter4K-S & Inter4K-L \\
\midrule
Ours (full)  & 0.9 & 4.6GB  & 34.16  & 30.45  & 29.29  & 25.16\\
w/o finetuning projection vectors  & 0.9  & 4.6GB  & 33.92   & 29.46  & 28.34 & 24.43 \\
w/o backward flow ($F_{t \rightarrow 0}$ \& $F_{t \rightarrow 1}$)   & 0.9  & 4.6GB  & 33.95  & 30.13  & 28.81 & 24.77\\
w/o $\mathcal{L}_{warp}$  &  0.9 & 4.6GB  & 33.99  & 30.25  & 28.97 & 24.96  \\
w/o $T$ scaling    &  0.9 & 4.6GB  & 34.13  & 30.41  & 29.20 &  25.09 \\
with synthesis (from~\citep{Sim:2021:XVF})   & 2.6 & 9.5GB  & 34.20  & 30.59 &  29.12 &   25.07 \\
\bottomrule
\end{tabularx}
\end{center}
    \vspace{-.5em}
\caption{\textbf{Ablation study.} Results from individually adding or removing one of the options from our full method. The largest benefit is obtained by finetuning the projection vectors.}
\label{table:ablation}
\end{table}


\subsection{Method analysis}
\paragraph{Efficiency.}
The focus of our method is to reduce the complexity of the network and the required memory. To select an ideal block size $d$ and the number of projection dimensions $k$, we analyze the general compression capability of block-based PCA by computing the basis from a single image of $512\times 512$, taking a fraction of it 
to reconstruct the images in X-Test~\citep{Sim:2021:XVF}, and averaging the reconstruction PSNR. For a targeted compression rate of $\nicefrac{1}{r}=\nicefrac{k}{d^2}=\nicefrac{1}{4}$, \cref{fig:reconstruction_error} shows that a block size of $d=8$ or $16$ has the best reconstruction accuracy. Because the compressed image representation is directly used for the flow computation and, therefore, a higher image resolution is preferable, we choose $d=8$ and thus $k=\nicefrac{d^2}{r}=\nicefrac{8^2}{4}=16$.
\cref{fig:comparison_features} shows the comparison of reconstructed images when using the original $k$ PCA vectors and after finetuning with our \fLDR{} module. The finetuned vectors increase the contrast, highlighting features that intuitively seem suitable for flow estimation in the context of frame interpolation.
The efficiency in feature representation using \fLDR{} becomes evident when looking at the remaining trainable parameters of our framework. 
In \cref{table:componenents} we report the memory usage, number of trainable parameters, and inference time \wrt the individual modules. 
Our flow network only has $0.73$ million (M) parameters, while XVFI~\citep{Sim:2021:XVF} needs $2.5$M for flow estimation plus an additional $1$M for flow refinement. By replacing the synthesis network of XVFI~\citep{Sim:2021:XVF} with weighted averaging (our fusion module), we save another $1.7$M parameters. 
Estimating optical flow with a pretrained PWC-Net~\citep{Sun:2018:PWC} requires 90ms at half the image resolution (to fit on our GPU), 9.4M trainable parameters, and 5.1GB memory. Visualizations of the optical flow are provided in the supplemental.
Our inference time is dominated by our warping module, designed to reduce the number of trainable parameters.\footnote{One trainable parameter is needed to compute the importance metric for forward warping as described in \cite{Niklaus:2020:SSV}.}

\paragraph{Ablation design choices.}
In \cref{table:ablation} we analyze the effect of our design choices. Finetuning the projection vectors is crucial and leads to a PSNR improvement of up to 1dB.
The variance of the final image quality (PSNR) for selecting different images for initialization of the projection dimensions is for Xiph-4K and X-Test only 0.02dB and 0.05dB, respectively. The Inter4K dataset has a larger variability in images, leading to a slighter higher variance of around 0.14dB. The powerful synthesis network only gives a slight positive effect in accuracy on one dataset, which is not proportional to the overhead it adds. The remaining design choices have a positive effect on the PSNR value while not adding any notable overhead.
\setlength{\tabcolsep}{2pt}
\begin{figure}[t]
\centering
\footnotesize
\begin{tabularx}{\linewidth}{@{}*5{>{\centering\arraybackslash}X}@{}}
\includegraphics[width=\linewidth]{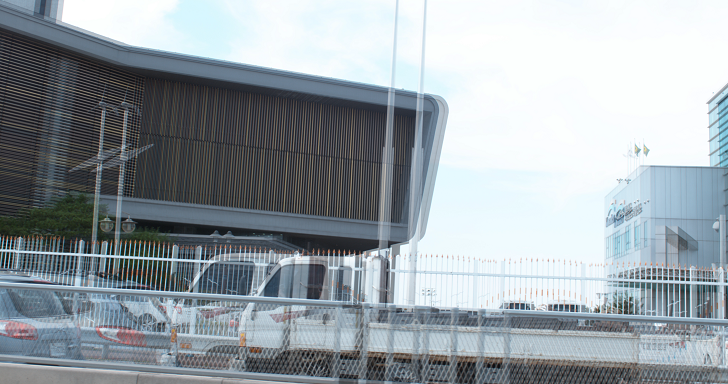} &
\includegraphics[width=\linewidth]{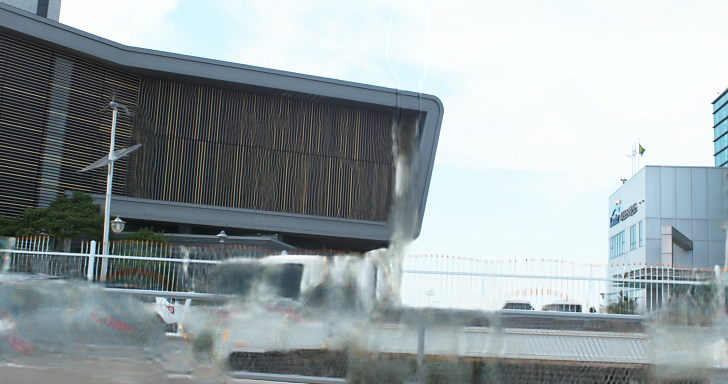} &
\includegraphics[width=\linewidth]{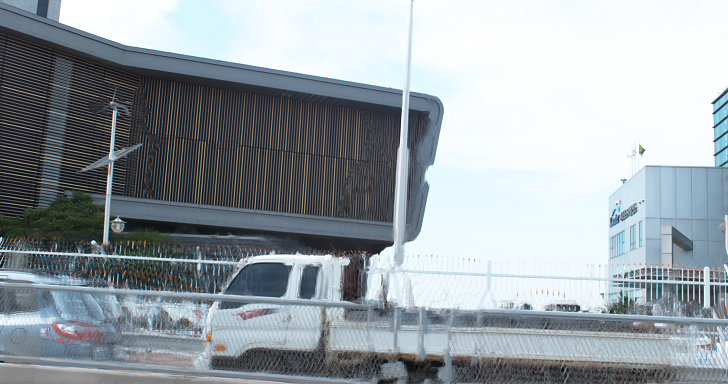} &
\includegraphics[width=\linewidth]{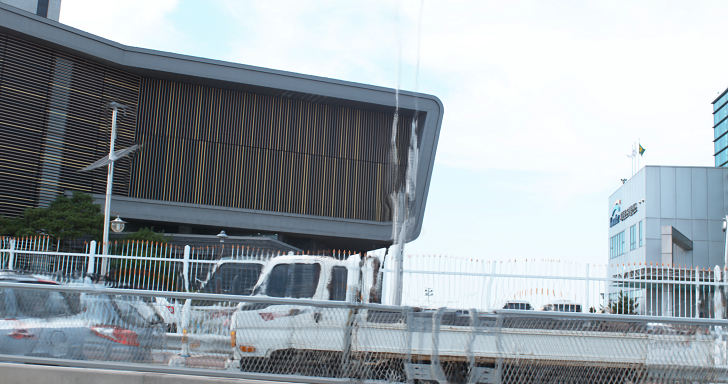} &
\includegraphics[width=\linewidth]{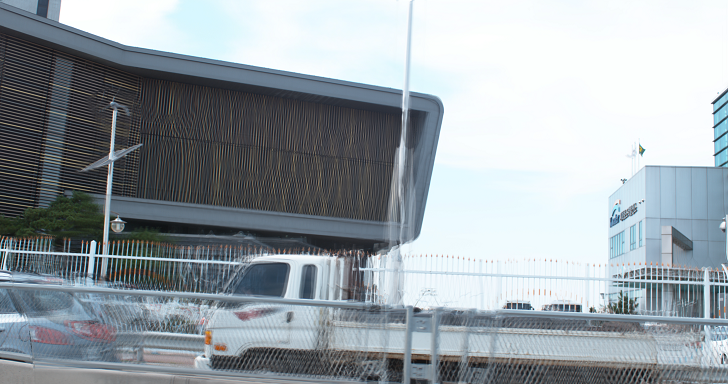} \\
\includegraphics[width=\linewidth]{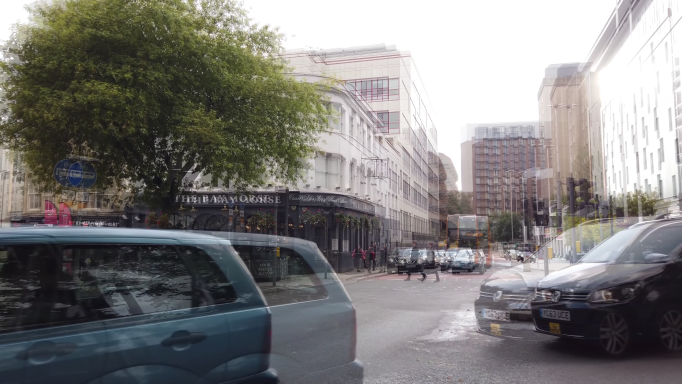} &
\includegraphics[width=\linewidth]{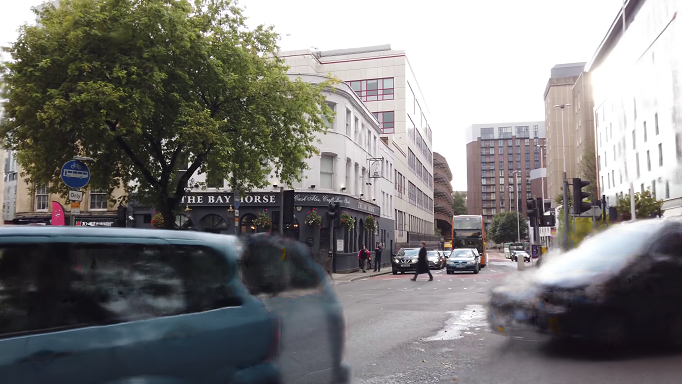} &
\includegraphics[width=\linewidth]{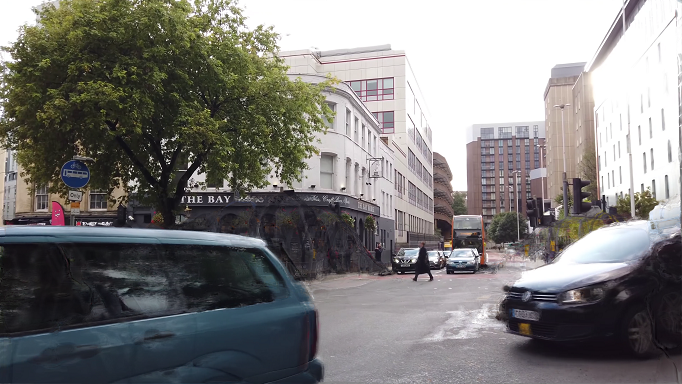} &
\includegraphics[width=\linewidth]{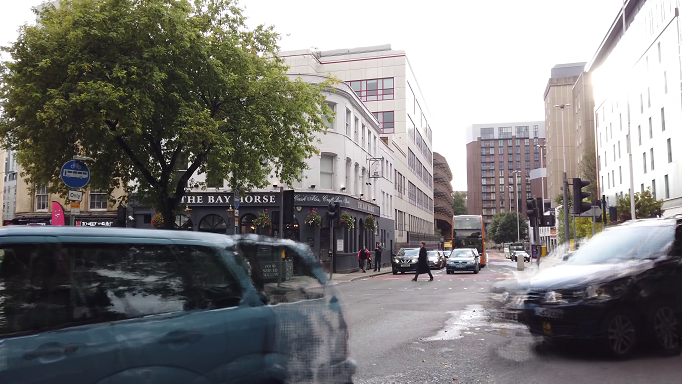} &
\includegraphics[width=\linewidth]{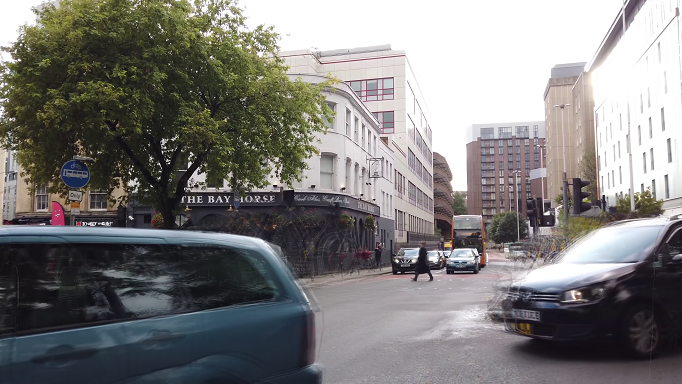} \\
\includegraphics[width=\linewidth]{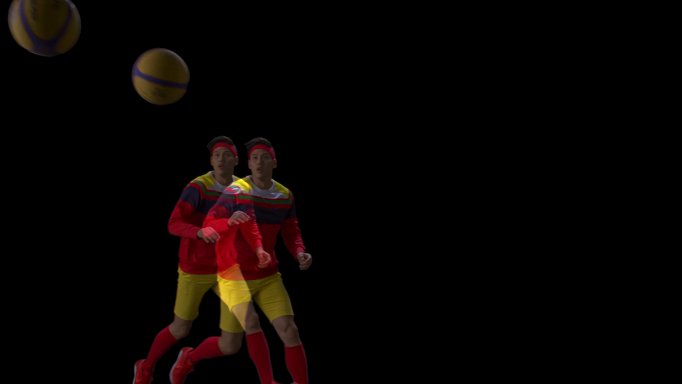} &
\includegraphics[width=\linewidth]{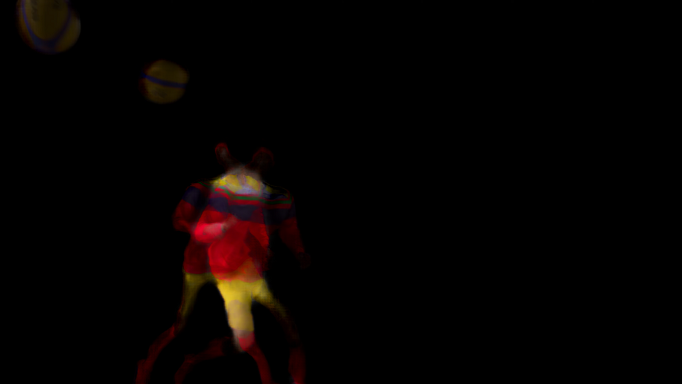} &
\includegraphics[width=\linewidth]{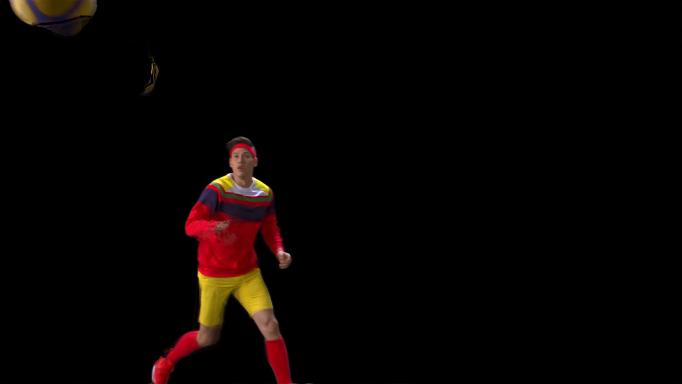} &
\includegraphics[width=\linewidth]{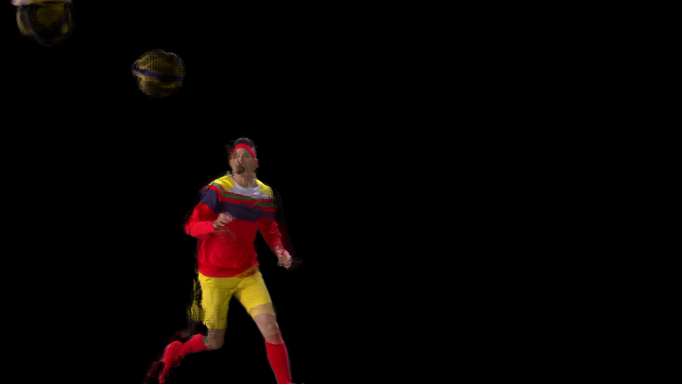} &
\includegraphics[width=\linewidth]{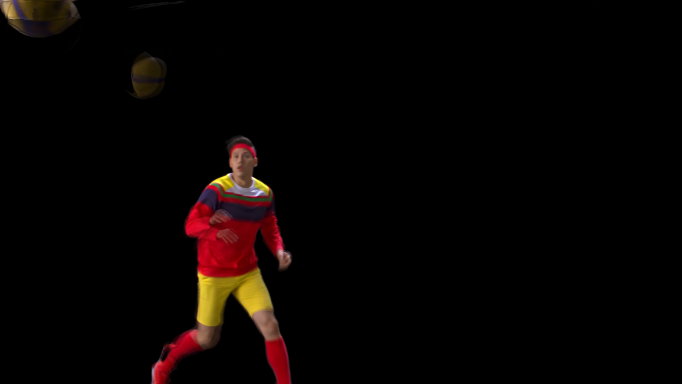} \\
\includegraphics[width=\linewidth]{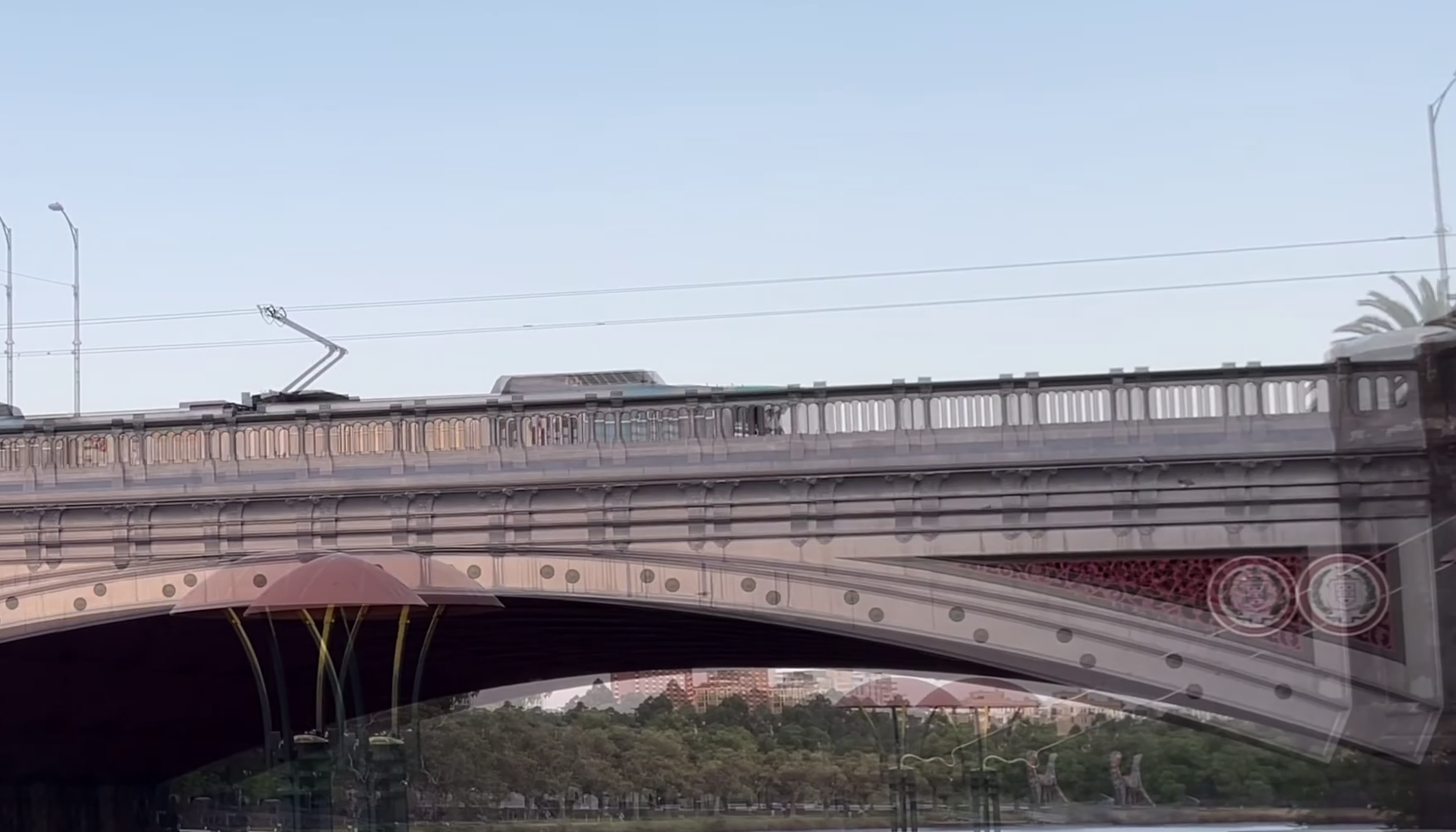} &
\includegraphics[width=\linewidth]{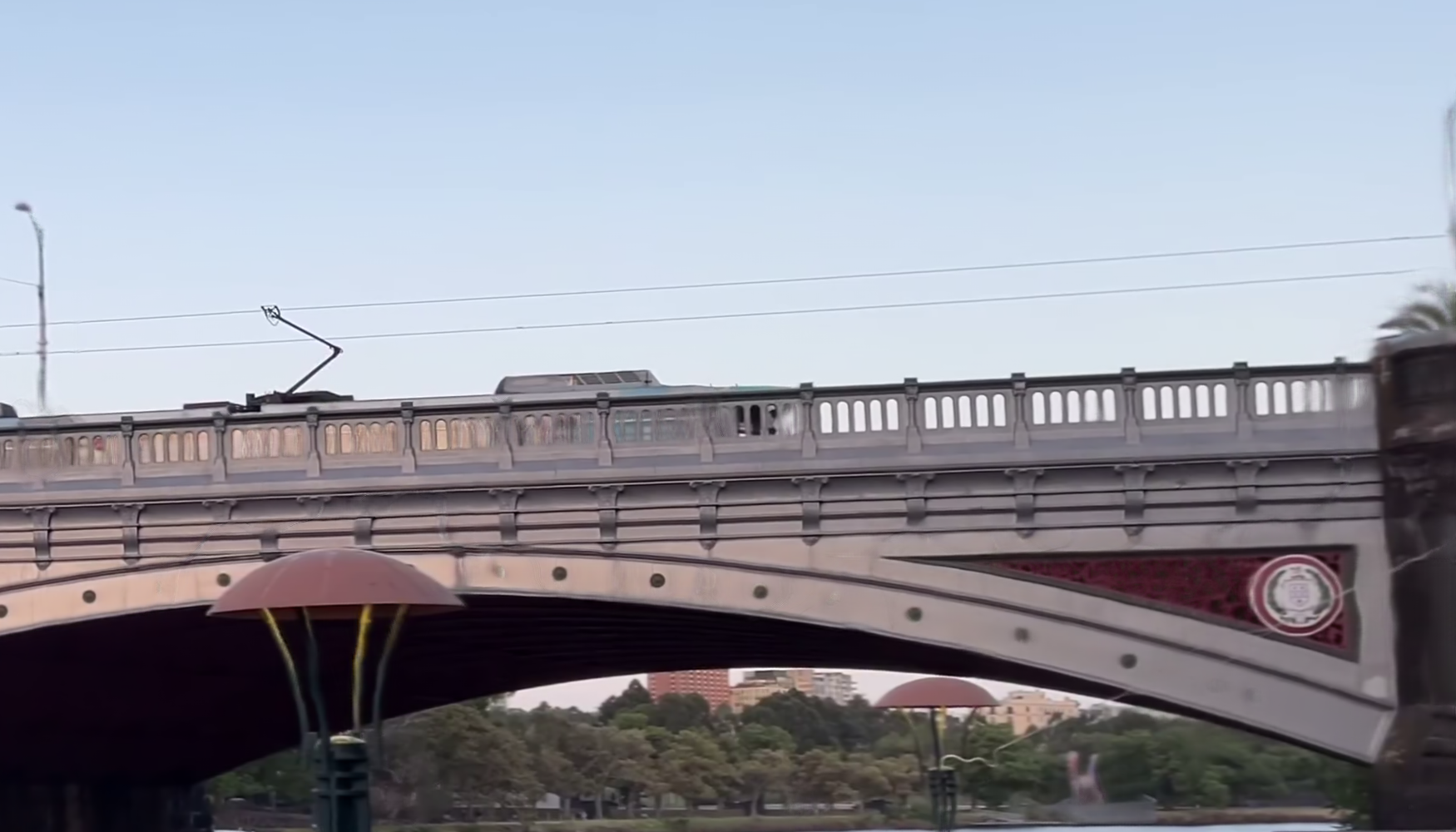} &
\includegraphics[width=\linewidth]{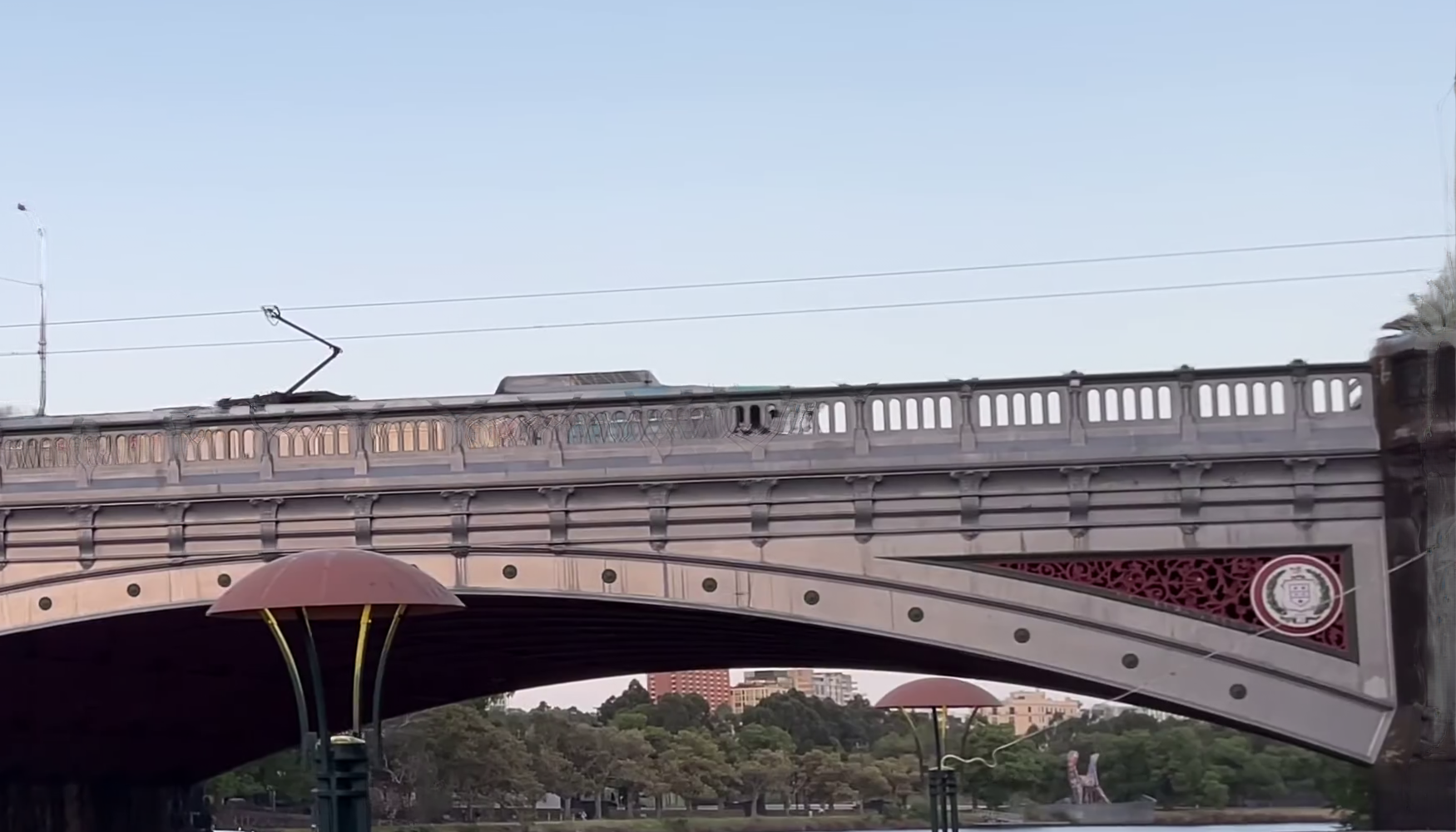} &
\includegraphics[width=\linewidth]{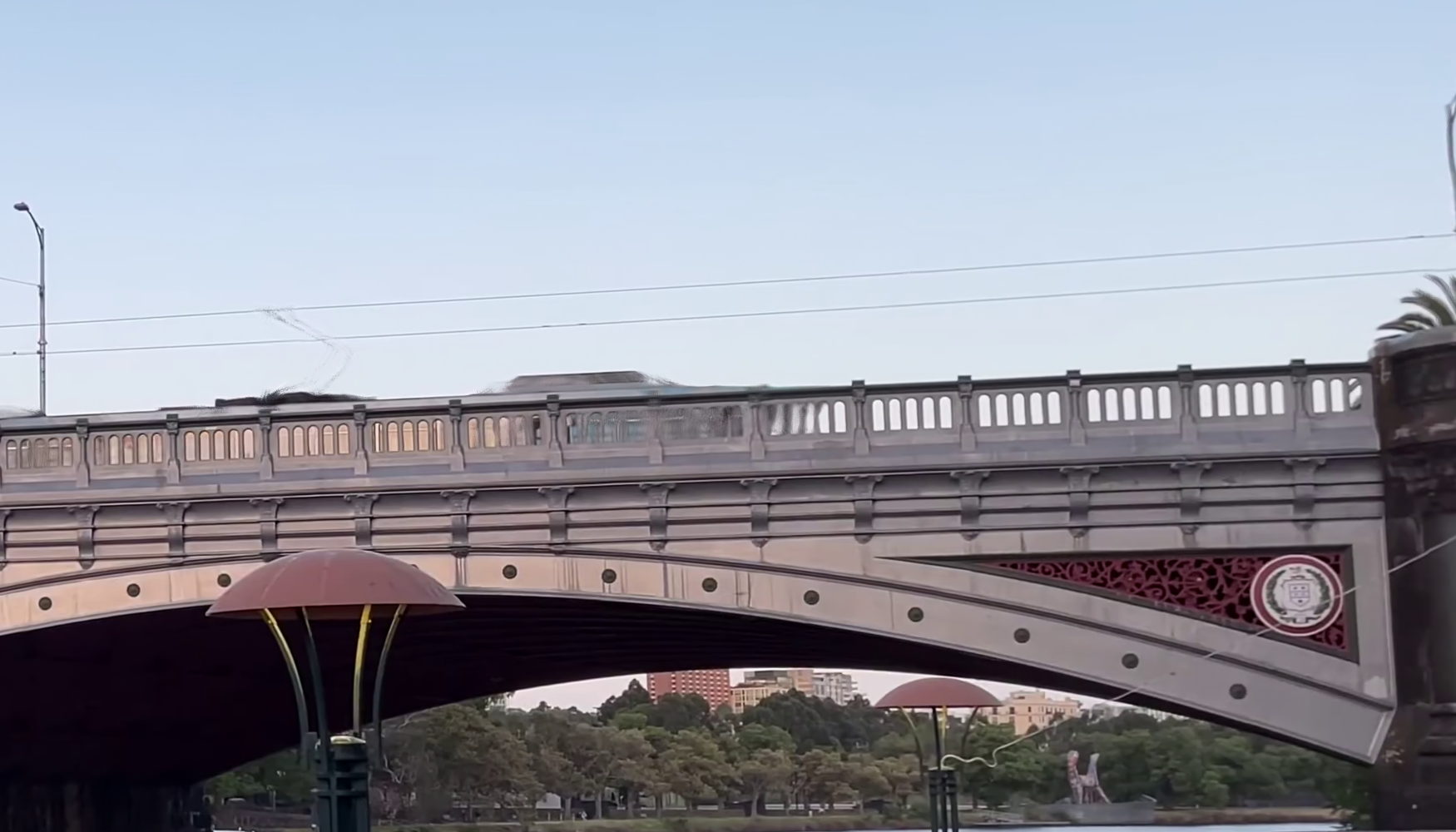} &
\includegraphics[width=\linewidth]{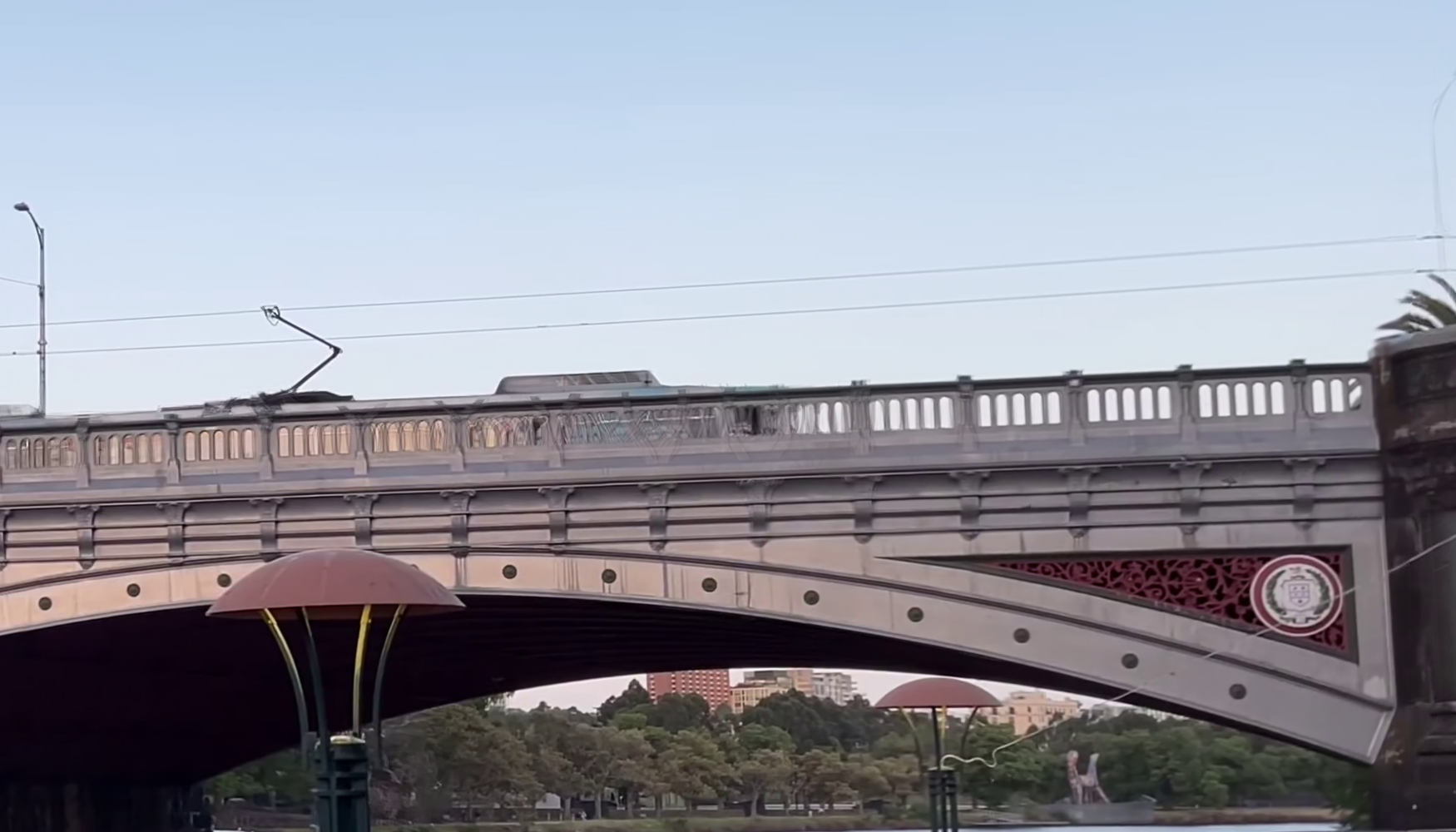} \\
(a) Overlaid inputs  & (b) RIFE$_m$~\citep{Huang:2020:RIF} & (c) XVFI~\citep{Sim:2021:XVF} & (d) M2M-PWC~\citep{Hu:2022:MMS} & (e) Ours\\
\end{tabularx}
    \vspace{-1em}
    \caption{\textbf{Qualitative examples.} Our method can handle large motion in 4K images better. From top to bottom, the images are taken from X-Test~\citep{Sim:2021:XVF} (\textit{top}), 
    Inter4K-S~\citep{Stergiou:2021:APE} (\textit{middle}, two examples), and Inter4k-L~\citep{Stergiou:2021:APE} (\textit{bottom}). The last row shows a cropped part of the original result. For Rife$_m$, we take the model trained on X-Train, giving better results on these datasets.}
    \label{fig:result_visual}
    \vspace{-1em}
\end{figure}
%
\paragraph{Limitations.}
Our framework has been explicitly and carefully designed for high-resolution video. 
Before flow computation, we compress the frames using linear dimensionality reduction on $8\times 8$ blocks. For this reason, flow can only be computed using $\nicefrac{1}{8}$ of the original image resolution as input. 
This makes our approach
perform comparatively worse on smaller images such as the common Vimeo90K~\citep{Xue:2019:VET} test set, 
 as the input to the flow estimation is not fine-grained enough \wrt to the full image resolution. Evaluating our method (trained on X-Train) directly on Vimeo-90k, we obtain a PSNR of 33.03dB using 1.7GB of memory. For small images, where memory constraints are less of an issue, better accuracy can be obtained with more memory intensive approaches; \eg, SoftSplat\citep{Niklaus:2020:SSV} achieves a PSNR of 36.10dB.

\section{Conclusion}
We present a method for video frame interpolation in high-resolution videos, which uses linear dimensionality reduction and a lightweight neural network architecture to boost the efficiency. By compressing images first with our finetuned projection vectors, we can significantly reduce the number of trainable parameters and the overall memory requirements in the subsequent network modules. 
We show that we can achieve highly competitive results on various 4K datasets with only a fraction of trainable parameters and low memory footprint. 

\section*{Acknowledgments and disclosure of funding}
This project has received funding from the
European Research Council (ERC) under the European Union’s Horizon 2020 research and innovation programme (grant agreement No.~866008).
The project has also been supported in part by the State of Hesse through the cluster project ``The Third Wave of Artificial Intelligence (3AI)''.

\bibliography{bibtex/long, mybib}

\newpage
\begin{appendices}
\section{Overview}
This appendix provides additional details related to training and the architectural setup for reproducibility purposes, which were omitted in the main paper due to space limitations. We further report evaluations with additional metrics and visualizations of the computed flow. The supplemental video\footnote{Available at \url{ https://github.com/visinf/fldr-vfi}.} contains additional, temporal visual results and comparisons.

\section{Training Details}
We give here the full formulas for the used loss functions. The total loss $\mathcal{L}_{total}$ in \cref{eqn:total_loss} consists of
\begin{align}
\mathcal{L}_{recon} &= \sum_{s=0}^{S} \| \hat{I}_{t}^{s} - I_{t}^{s} \|_1  \\
\mathcal{L}_{smooth} &= \sum_{(i,j)\in\{(0,1), (1,0)\}} \exp\left(-e^2 \sum_c  (\nabla_x I_{i_c})^2\right)^\mathsf{T} \cdot |\nabla_x F_{i \rightarrow j}^0| \\
\mathcal{L}_{warp} &= \left( \| \overrightarrow{\omega}(I_0^0,F_{0 \rightarrow 1}) - I_1^0  \|_1 +  \| \overrightarrow{\omega}(I_1^0,F_{1\rightarrow 0})- I_0^0 \|_1 \right) \; ,
\end{align}
where $s$, $c$, $e$, $x$, $\lambda_{\cdot}$, and $\overrightarrow{\omega}(\cdot, \cdot)$ define the scale, the channel index, an edge weighting factor, a spatial coordinate, the weighting factors, and the forward warping function, respectively. The values for the hyperparameters are given in \cref{tab:hyperparameters}.

\begin{table}[h]
\footnotesize
    \centering
    \begin{tabularx}{0.5\linewidth}{@{}*4{>{\centering\arraybackslash}X}@{}}
    \toprule
         $S_{train}$  & $e$  &  $\lambda_{smooth}$ & $\lambda_{warp}$ \\
         \midrule
         3 &  150 & 0.125 & 0.5  \\
    \bottomrule
    \end{tabularx}
    \smallskip
    \caption{Overview of the used hyperparameters.}
    \label{tab:hyperparameters}
\end{table}

As mentioned in the occlusion estimation in \cref{sec:architecture}, we apply the softmax function along the last dimension of the weighting map, creating a probability distribution for every output pixel. Using the softmax function allows for temperature scaling~\citep{Guo:2017:CMN}, for which we divide the estimated occlusion map $M$ by the scalar temperature parameter $T$ before taking the softmax. This allows to soften the distribution or make it more peaked, depending on whether $T$ its greater or smaller than $1$. After training the full pipeline with $\mathcal{L}_{total}$ and $T=1$, we additionally end-to-end optimize for $T$ while keeping all other parameters fixed. We finetune $T$ with a learning rate of $10^{-3}$ using the mean-squared-error of the image reconstruction loss for 10 epochs.

\section{Network Architecture}
\cref{tab:flow_coarse,tab:flow_all} list the details of our flow estimation network for the coarsest scale and all higher levels, respectively.
After each convolutional layer, ReLU nonlinearities~\citep{Nair:2010:RLU} are applied except for conv2d\_7 in \cref{tab:flow_coarse}, and conv2d\_3$_i$ and conv2d\_8 in \cref{tab:flow_all}.
The kernel size of each convolutional layer 
is $3 \times 3$.  Layer conv2d\_1 and conv2d\_2 in \cref{tab:flow_coarse,tab:flow_all} share their parameters, as well as conv2d\_3$_0$ and conv2d\_3$_1$ in \cref{tab:flow_all}.
$\textrm{fLDR}^s = [\textrm{fLDR}^s_0$, $\textrm{fLDR}^s_1]$ is the result of our finetuned linear dimensionality reduction (fLDR) of image $I_0$ and $I_1$ at scale $s$, \ie $[\Tilde{I}^s_0, \Tilde{I}^s_1]$.

\cref{tab:flow_all} shows the architecture of our flow estimation network for $s<S$, with $\textrm{feat}_0$ and $\textrm{feat}_1$  defined as follows:
\begin{align}
 \textrm{feat}_0 : \ & \left[(\textrm{fLDR}_0^s + \textrm{conv2d}\_2_0) , \overrightarrow{\omega}(\textrm{fLDR}_0^s + \textrm{conv2d}\_2_0,\mathit{up}(F_{0 \rightarrow 1}^{s+1})) \right]
         \\  
            \textrm{feat}_1 : \ & \left[(\textrm{fLDR}^s_1 + \textrm{conv2d}\_2_1) , \overrightarrow{\omega}(\textrm{fLDR}^s_1 + \textrm{conv2d}\_2_1,\mathit{up}(F_{1\rightarrow 0}^{s+1}))\right] \ ,
\end{align}
where $F_{0 \rightarrow 1}^{s+1}$ and $F_{1 \rightarrow 0}^{s+1}$ are the estimated, bidirectional flows from the previous,  coarser scale and $\textrm{conv2d}\_2_0$ as well as $\textrm{conv2d}\_2_1$ are the outputs of the convolutional layer conv2d\_2, split along the channel dimension, such that they represent the features of input image  $I_0$ and $I_1$, respectively. The flow has been upscaled ($\mathit{up}(\cdot)$) bilinearly. 

The configuration of the occlusion estimation network is listed in \cref{tab:occlusion}.
After each convolutional layer, ReLU nonlinearities~\citep{Nair:2010:RLU} are applied except for layer dec\_3.
\begin{table}[t]
    \centering
\footnotesize
\begin{tabular}{@{}lllll@{}}
        \toprule
\multicolumn{1}{c}{Layer name}& \multicolumn{1}{c}{Input}    &\multicolumn{1}{c}{\# Channels In/Out} &\multicolumn{1}{c}{Reuse weights}
                    \\ \midrule
                    
        \multicolumn{1}{c}{conv2d\_1} & \multicolumn{1}{c}{$\textrm{fLDR}^s$}                  & \multicolumn{1}{c}{$96/96$} & \multicolumn{1}{c}{False}  \\ 
        
        \multicolumn{1}{c}{conv2d\_2} & \multicolumn{1}{c}{conv2d\_1}       & \multicolumn{1}{c}{$96/96$} & \multicolumn{1}{c}{False} \\ 
        
        \multicolumn{1}{c}{conv2d\_3} & \multicolumn{1}{c}{conv2d\_2 + $\textrm{fLDR}^s$}                 & \multicolumn{1}{c}{$96/96$}  & \multicolumn{1}{c}{False}  \\

        \multicolumn{1}{c}{conv2d\_4} & \multicolumn{1}{c}{conv2d\_3}                &   \multicolumn{1}{c}{$96/96$}  & \multicolumn{1}{c}{False}  \\ 
        
        \multicolumn{1}{c}{conv2d\_5} & \multicolumn{1}{c}{conv2d\_4}                &   \multicolumn{1}{c}{$96/96$}  & \multicolumn{1}{c}{False}  \\ 
        
        \multicolumn{1}{c}{conv2d\_6} & \multicolumn{1}{c}{conv2d\_5}                &  \multicolumn{1}{c}{$96/48$}  & \multicolumn{1}{c}{False}  \\ 
        
         \multicolumn{1}{c}{conv2d\_7} & \multicolumn{1}{c}{conv2d\_6}                &   \multicolumn{1}{c}{$48/4$}  & \multicolumn{1}{c}{False}  \\

         \bottomrule
        \end{tabular}
        \smallskip
        \caption{Layers of the flow estimation network at the coarsest scale $s=S$.  The column ``Reuse weights'' indicates if the parameters of a layer are reused from another layer. The output of the last layer is the bidirectional flow $[F_{0\rightarrow 1}^{S},F_{1\rightarrow 0}^{S}]$.
        }
        \label{tab:flow_coarse}
\end{table}

\begin{table}[t]
    \centering
\footnotesize
        \begin{tabular}{@{}lllll@{}}
        \toprule
\multicolumn{1}{c}{Layer name}& \multicolumn{1}{c}{Input}    &\multicolumn{1}{c}{\# Channels In/Out} &\multicolumn{1}{c}{Reuse weights}
                    \\ \midrule
                    
        \multicolumn{1}{c}{conv2d\_1} & \multicolumn{1}{c}{$\textrm{fLDR}^s$}                  & \multicolumn{1}{c}{$96/96$} & \multicolumn{1}{c}{conv2d\_1 (\cref{tab:flow_coarse})}  \\ 
        
        \multicolumn{1}{c}{conv2d\_2} & \multicolumn{1}{c}{conv2d\_1}       & \multicolumn{1}{c}{$96/96$} & \multicolumn{1}{c}{conv2d\_2 (\cref{tab:flow_coarse})} \\ 
        
        \multicolumn{1}{c}{conv2d\_3$_0$} & \multicolumn{1}{c}{$\textrm{feat}_0$}                 & \multicolumn{1}{c}{$96/48$}  & \multicolumn{1}{c}{False}  \\ 
        \multicolumn{1}{c}{conv2d\_3$_1$} & \multicolumn{1}{c}{$\textrm{feat}_1$}               
        & \multicolumn{1}{c}{$96/48$}  & \multicolumn{1}{c}{conv2d\_3$_0$}  \\ 
        
        \multicolumn{1}{c}{conv2d\_4} & \multicolumn{1}{c}{[$\textrm{conv2d\_3}_0, \textrm{conv2d\_3}_1, \mathit{up}(F^{s+1})$]}                &   \multicolumn{1}{c}{$100/96$}  & \multicolumn{1}{c}{False}  \\ 
        
        \multicolumn{1}{c}{conv2d\_5} & \multicolumn{1}{c}{conv2d\_4}                &   \multicolumn{1}{c}{$96/96$}  & \multicolumn{1}{c}{False}  \\ 
        
        \multicolumn{1}{c}{conv2d\_6} & \multicolumn{1}{c}{conv2d\_5}                &  \multicolumn{1}{c}{$96/48$}  & \multicolumn{1}{c}{False}  \\ 
        
         \multicolumn{1}{c}{conv2d\_7} & \multicolumn{1}{c}{conv2d\_6}                &   \multicolumn{1}{c}{$48/48$}  & \multicolumn{1}{c}{False}  \\ 
        
        \multicolumn{1}{c}{conv2d\_8} & \multicolumn{1}{c}{conv2d\_7}                &   \multicolumn{1}{c}{$48/4$}  & \multicolumn{1}{c}{False}  \\
        
         \bottomrule
        \end{tabular}
        \smallskip
        \caption{Layers of the flow estimation network at all scales except the coarsest, \ie $s<S$.  
        The column ``Reuse weights'' indicates if the parameters of a layer are reused from another layer. The output of the last layer is added to the upsampled flow $F^{s+1}$ of the previous scale, which results in the final bidirectional flow $F^s=[F_{0\rightarrow 1}^{s},F_{1\rightarrow 0}^{s}]$ of scale s.
        }
        \label{tab:flow_all}
\end{table}

\begin{table}[t]
    \centering
\footnotesize
        \begin{tabular}{@{}llll@{}}
        \toprule
\multicolumn{1}{c}{Layer name} & \multicolumn{1}{c}{Filter Size}   &\multicolumn{1}{c}{\# Input Filters}&\multicolumn{1}{c}{\# Output Filters}       
                    \\ \midrule
        
        \multicolumn{1}{c}{enc\_0}                 & \multicolumn{1}{c}{$4\times 4$}  & \multicolumn{1}{c}{$26$} & \multicolumn{1}{c}{$16$} \\             
        \multicolumn{1}{c}{enc\_1}                 & \multicolumn{1}{c}{$4\times 4$}  & \multicolumn{1}{c}{$16$} & \multicolumn{1}{c}{$32$} \\ 
        
        \multicolumn{1}{c}{enc\_2}                 & \multicolumn{1}{c}{$4\times 4$}  & \multicolumn{1}{c}{$32$}  & \multicolumn{1}{c}{$64$} \\ 
         
          \multicolumn{1}{c}{dec\_0}                 & \multicolumn{1}{c}{$3\times 3$}  & \multicolumn{1}{c}{64}  & \multicolumn{1}{c}{$64$} \\ 
          
       \multicolumn{1}{c}{dec\_1}                 & \multicolumn{1}{c}{$3\times 3$}  & \multicolumn{1}{c}{$64 + 32$}  & \multicolumn{1}{c}{$32$} \\          
       \multicolumn{1}{c}{dec\_2}                 & \multicolumn{1}{c}{$3\times 3$}  & \multicolumn{1}{c}{$32 + 16$}  & \multicolumn{1}{c}{$16$} \\    
       \multicolumn{1}{c}{dec\_3}                 & \multicolumn{1}{c}{$3\times 3$}  & \multicolumn{1}{c}{$16$}  & \multicolumn{1}{c}{$6$} \\

         \bottomrule
        \end{tabular}
        \smallskip
        \caption{Layers of the occlusion estimation network. The output of enc\_2 is additionally fed into dec\_1. The output of enc\_1 is additionally fed into dec\_2. The outputs of dec\_0, dec\_1, dec\_2, and dec\_3 are upscaled with nearest neighbor upsampling and a scale factor of 2 before feeding them into the respective next layer.}
        \label{tab:occlusion}
\end{table}

\section{Quantitative Results}
In \cref{table:additional_metrics_4K} we provide, in addition to the PSNR values in  \cref{table:results_4K}, a quantitative analysis with SSIM~\citep{Wang:2004:IQA}, LPIPS~\cite{Zhang:2018:UED}, and inference time, where possible. We can only measure the inference time for models where we have the code and which are running on our Nvidia 3080Ti (12GB) GPU. Unfortunately, XVFI with $S=5$ scales run out of memory on our GPU with 12 GB and the error metric computation has been computed on CPU. We, therefore, provide the inference time only for $S=3$ ($^\ddagger$). We did not optimize our method for inference time. Nevertheless, we obtain inference times  comparable to most of the other methods. M2M-PWC downscales the input image first by a factor of $\nicefrac{1}{16}$, leading to faster inference.
\setlength{\tabcolsep}{4pt}
\begin{table}[t]
\footnotesize
    \centering
\begin{tabularx}{\linewidth}{@{}l*5{>{\centering\arraybackslash}X}@{}}
\toprule
& Xiph-4K
& X-Test 
& Inter4K-S & Inter4K-L & Inference (in s/f)\\
\midrule
M2M-PWC~\cite{Hu:2022:MMS}  & \textbf{0.949}/0.219 & \textbf{0.914}/\underline{0.086} & \textbf{0.942}/\textbf{0.076} & \underline{0.883}/\underline{0.145} & \textbf{0.21}   \\
\midrule
RIFE$_{m}^\diamond$~\cite{Huang:2020:RIF}  & 0.910/\underline{0.171} & 0.793/0.227 & 0.894/0.117 & 0.826/0.196 & \underline{0.40} \\
RIFE$_{m}$~\cite{Huang:2020:RIF}  & 0.904/0.228 & 0.751/0.260 & 0.893/0.123 & 0.829/0.197 & \underline{0.40}  \\
XVFI$^\diamond$ \cite{Sim:2021:XVF}  & 0.910/0.175 & \underline{0.874}/\textbf{0.085} & 0.915/0.085 & 0.850/\underline{0.145} & 0.66$^\ddagger$/ N/A    \\
Ours  & \underline{0.913}/\textbf{0.144} & 0.871/0.099 & \underline{0.917}/\underline{0.084} & \textbf{0.904}/\textbf{0.139} & 0.50$^\ddagger$/0.51 \\
\bottomrule
\end{tabularx}
\vspace{-0.5em}
\caption{Extension of \cref{table:results_4K} with (SSIM~\citep{Wang:2004:IQA}/LPIPS~\citep{Zhang:2018:UED}) and inference time in s for a 4K image ($2160\times 4096$) on a Nvidia 3080Ti GPU. 
}
\label{table:additional_metrics_4K}
\end{table}

\section{Optical Flow Visualizations}
\begin{figure}[t]
\centering
\footnotesize
\begin{tabularx}{\linewidth}{@{}*5{>{\centering\arraybackslash}X}@{}}
\includegraphics[width=\linewidth]{images/images_qualitive/xtest/middleframe.png} &
\includegraphics[width=\linewidth]{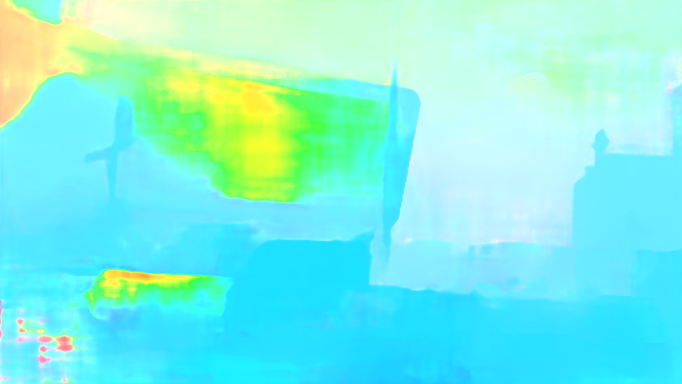}
 & \includegraphics[width=\linewidth]{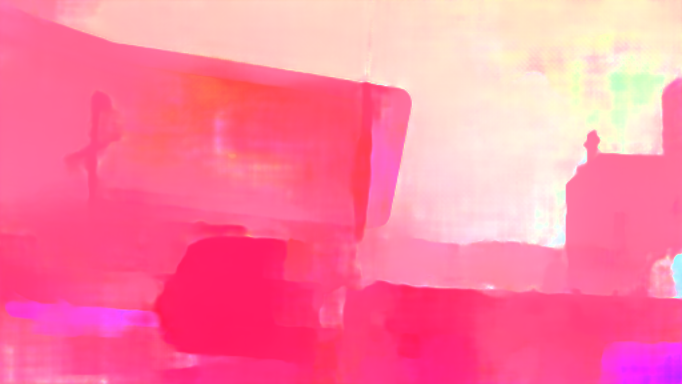}
 & \includegraphics[width=\linewidth]{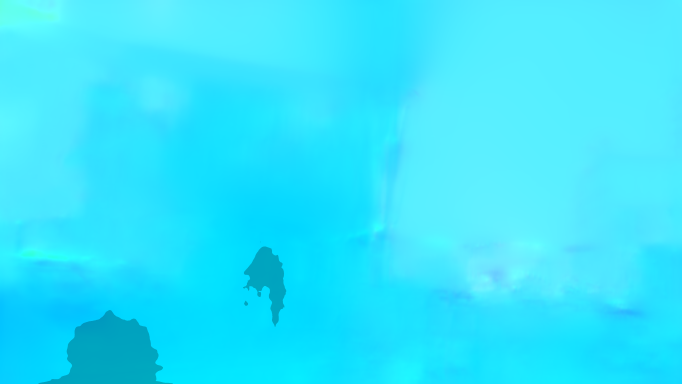}
 & \includegraphics[width=\linewidth]{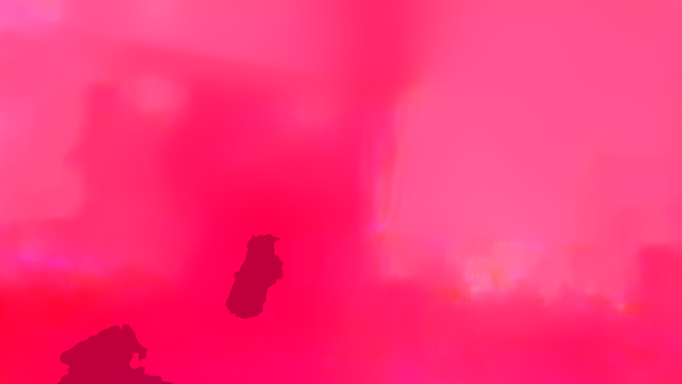}
\\
\includegraphics[width=\linewidth]{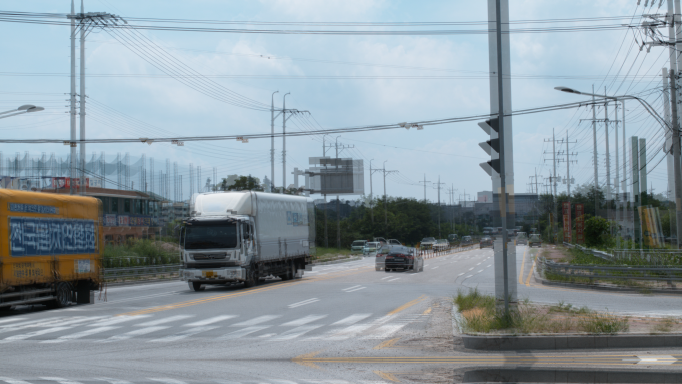} &
\includegraphics[width=\linewidth]{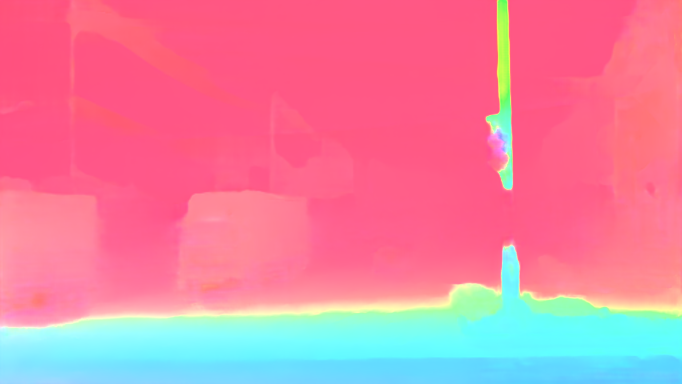}
 & \includegraphics[width=\linewidth]{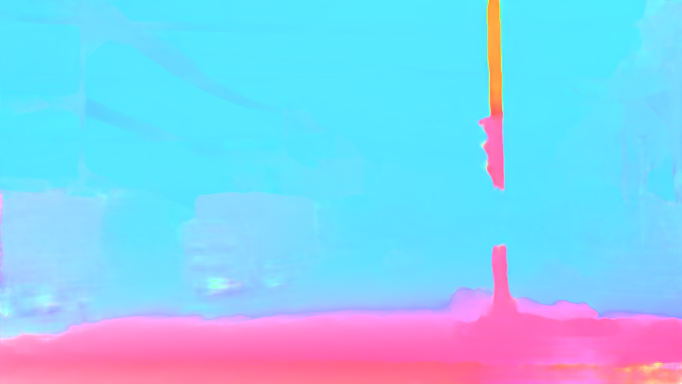}
 & \includegraphics[width=\linewidth]{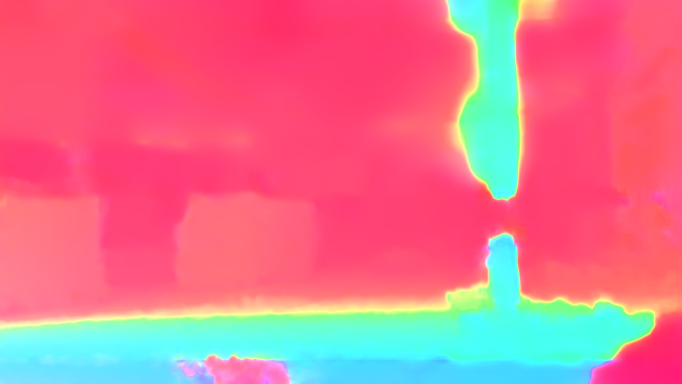}
 & \includegraphics[width=\linewidth]{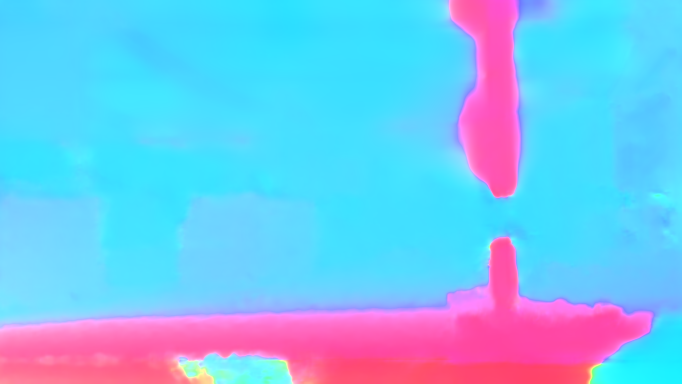}
\\

\includegraphics[width=\linewidth]{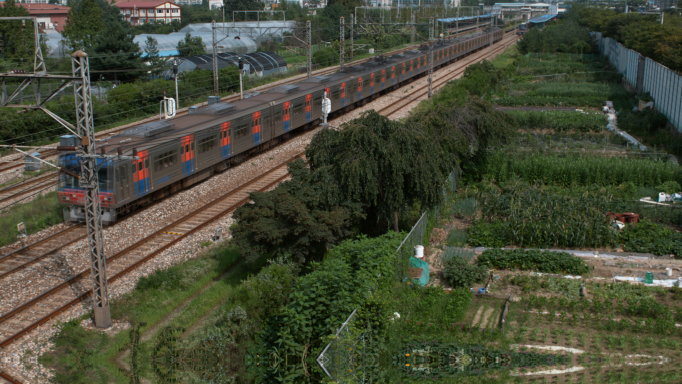} &
\includegraphics[width=\linewidth]{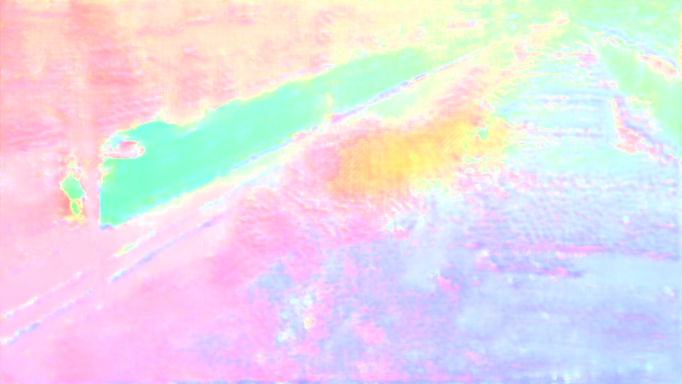}
 & \includegraphics[width=\linewidth]{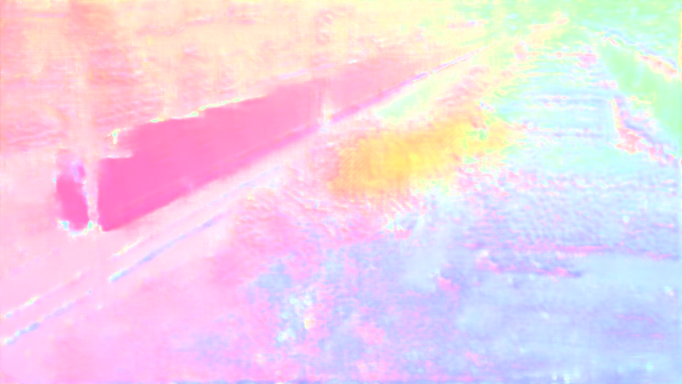}
 & \includegraphics[width=\linewidth]{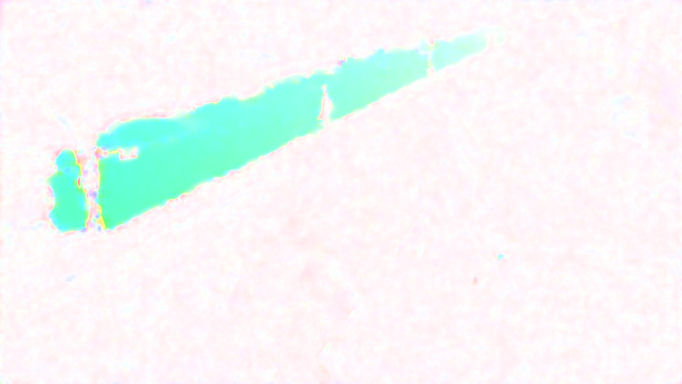}
 & \includegraphics[width=\linewidth]{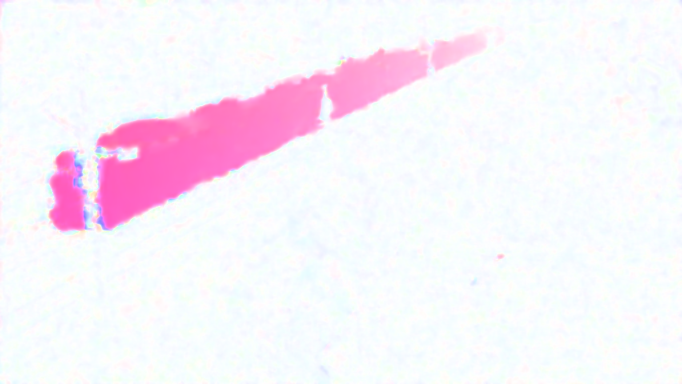}
\\
\includegraphics[width=\linewidth]{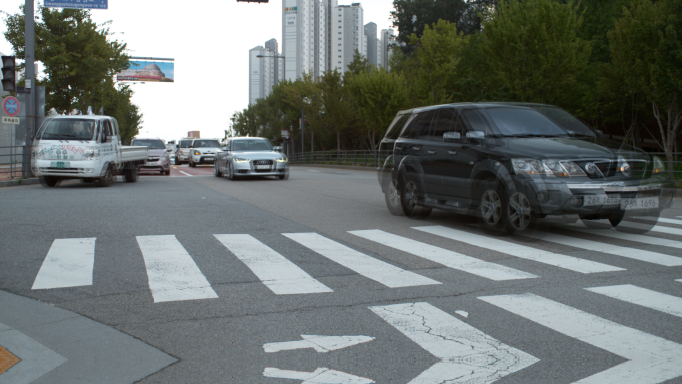} &
\includegraphics[width=\linewidth]{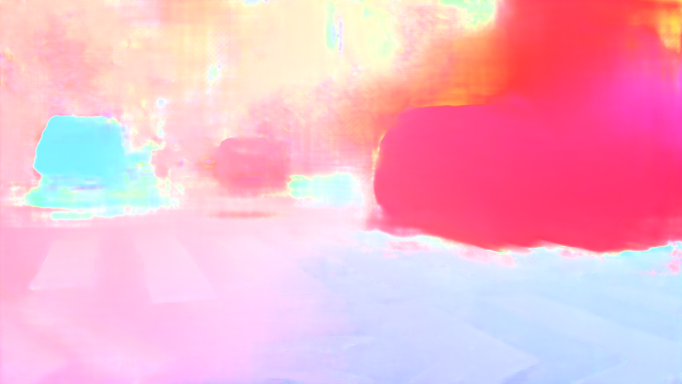}
 & \includegraphics[width=\linewidth]{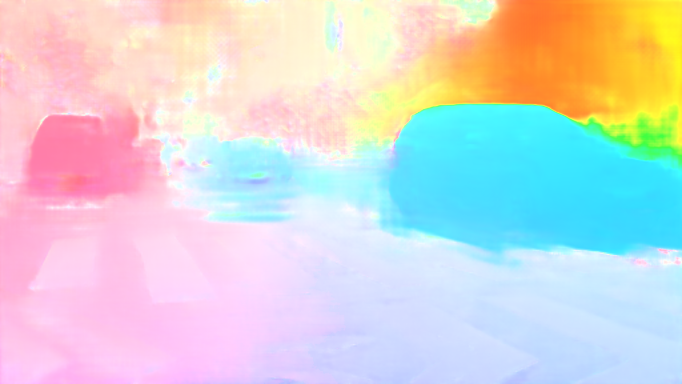}
 & \includegraphics[width=\linewidth]{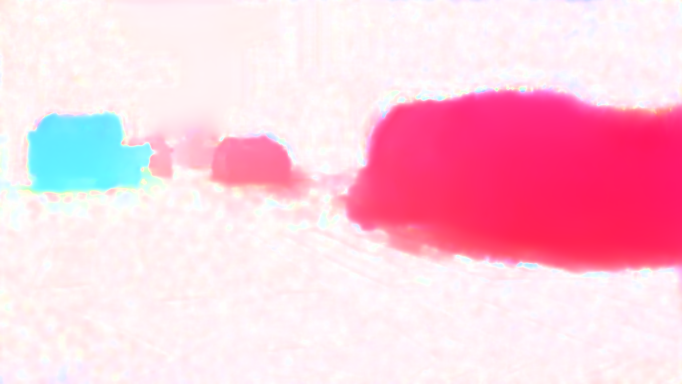}
 & \includegraphics[width=\linewidth]{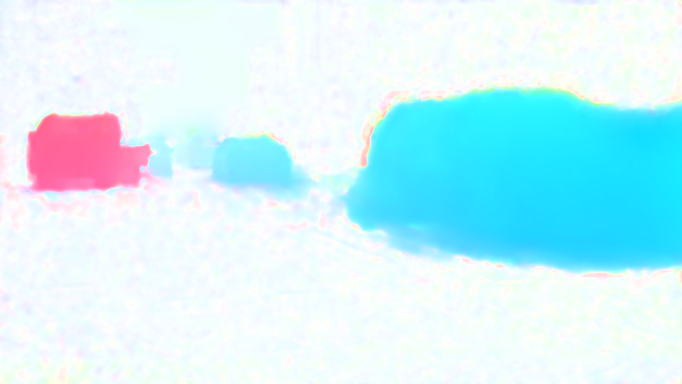}
\\
(a) Overlaid inputs  & (b) $F_{0 \rightarrow 1}$ \citep{Sun:2018:PWC} & (c) $F_{1 \rightarrow 0}$ \citep{Sun:2018:PWC} & (d) $F_{0 \rightarrow 1}$ (Ours) & (e) $F_{1 \rightarrow 0}$ (Ours)\\
\end{tabularx}
\vspace{-0.5em}
    \caption{\textbf{Flow visualization.} Comparison of computed flow between ours (smaller network, flow for frame interpolation) and the pretrained PWC-Net~\citep{Sun:2018:PWC}.}
    \label{fig:flow_visual}
    \vspace{-1em}
\end{figure}

In \cref{fig:flow_visual} we show some visualizations of our predicted flow in comparison to the computationally more expensive pretrained PWC-Net~\citep{Sun:2018:PWC} often used in frame interpolation methods~\citep{Niklaus:2020:SSV, Hu:2022:MMS}. However, the flow shown here, is computed with the original PWC-Net without yet finetuning for the task of frame interpolation as this depends on the used frame interpolation method.
\setlength{\tabcolsep}{2pt}
\end{appendices}

\end{document}